%% file: egpaper_for_review.tex
\documentclass[10pt,twocolumn,letterpaper]{article}
\usepackage{titling}
\usepackage{cvpr}
\usepackage{times}
\usepackage{epsfig}
\usepackage{graphicx}
\usepackage{array}
\usepackage{amsmath}
\usepackage{amsfonts}
\usepackage{multirow}
\usepackage{amssymb}
\usepackage{booktabs}
\usepackage{ragged2e}
\newcolumntype{P}[1]{>{\RaggedRight\hspace{0pt}}p{#1}}
\newcommand{\ra}[1]{\renewcommand{\arraystretch}{#1}}

\DeclareMathOperator*{\argmin}{arg\,min}

\usepackage{float}
\usepackage[dvipsnames]{xcolor}

\usepackage[pagebackref=true,breaklinks=true,letterpaper=true,colorlinks,bookmarks=false]{hyperref}

\cvprfinalcopy 

\def\cvprPaperID{****} 

\newcommand{\icg}[2]{\includegraphics[width=#1\linewidth]{#2}}

\newcommand{\comment}[1]{}

\newcommand\blfootnote[1]{%
  \begingroup
  \renewcommand\thefootnote{}\footnote{#1}%
  \addtocounter{footnote}{-1}%
  \endgroup
}

\ifcvprfinal\fi
\begin{document}

\title{Predicting Sharp and Accurate Occlusion Boundaries in \\Monocular Depth Estimation Using Displacement Fields}

\author{Micha\"el Ramamonjisoa$^{*} \quad\quad$ Yuming Du$^{*} \quad\quad$ Vincent Lepetit \\
  LIGM, IMAGINE, Ecole des Ponts, Univ Gustave Eiffel, CNRS, Marne-la-Vall\'ee France\\
  {\tt\small \{first.lastname\}@enpc.fr}
{\small\url{https://michaelramamonjisoa.github.io/projects/DisplacementFields}}}

\maketitle

\begin{abstract}
  
  Current methods for depth map prediction from monocular images tend to predict
  smooth, poorly  localized contours for  the occlusion boundaries in  the input
  image.   This is  unfortunate as  occlusion boundaries  are important  cues to
  recognize objects, and as  we show, may lead to a way  to discover new objects
  from scene  reconstruction.  To improve  predicted depth maps,  recent methods
  rely on various  forms of filtering or predict an  additive residual depth map
  to refine a  first estimate.  We instead  learn to predict, given  a depth map
  predicted  by some  reconstruction method,  a  2D displacement  field able  to
  re-sample pixels around the occlusion boundaries into sharper reconstructions.
  Our method can be applied to the  output of any depth estimation method and is
  fully  differentiable,  enabling  end-to-end  training.   For  evaluation,  we
  manually annotated  the occlusion  boundaries in  all the  images in  the test
  split of popular  NYUv2-Depth dataset. We show that our  approach improves the
  localization of occlusion boundaries  for all state-of-the-art monocular depth
  estimation   methods   that    we   could   evaluate~(\cite{Laina2016DeeperDP,
    FuCVPR18-DORN,  Eigen15,  Jiao2018LookDI}),   without  degrading  the  depth
  accuracy for the rest of the images.
    
\end{abstract}

\blfootnote{$^*$ Authors with equal contribution.}

\section{Introduction}
\input{intro}

\section{Related Work}
\input{related}

\section{Method}
\input{method}

\section{Experiments}
\input{experiments}

\section{Conclusion}
\input{conclusion}

\vspace{-5pt}
\section*{Acknowledgement}
\vspace{-5pt}
We thank Xuchong Qiu and Yang Xiao for their help in annotating the NYUv2-OC++ dataset. This project has received funding from the CHISTERA-IPALM project.	

\input{supplementary_materials}

{\small
  \bibliographystyle{ieee_fullname}
  \bibliography{short,vision,biblio}
}

\end{document}

%% file: intro.tex

Monocular depth  estimation~(MDE) aims at predicting  a depth map from  a single
input image. It attracts  many interests, as it can be  useful for many computer
vision tasks  and applications, such  as scene understanding,  robotic grasping,
augmented reality, etc.  Recently, many methods have been proposed to solve this
problem   using  Deep   Learning  approaches,   either  relying   on  supervised
learning~\cite{Eigen14,   Eigen15,  Laina2016DeeperDP,   FuCVPR18-DORN}  or   on
self-learning~\cite{Monodepth17,Xie2016Deep3DFA,PoggiTrinocular},    and   these
methods already often obtain very impressive results.

\begin{figure}[H]
	\begin{center}
		\begin{tabular}{>{\centering\arraybackslash}m{0.375\linewidth}
				>{\centering\arraybackslash}m{0.375\linewidth}}
			\icg{0.9}{}	&
			\icg{0.9}{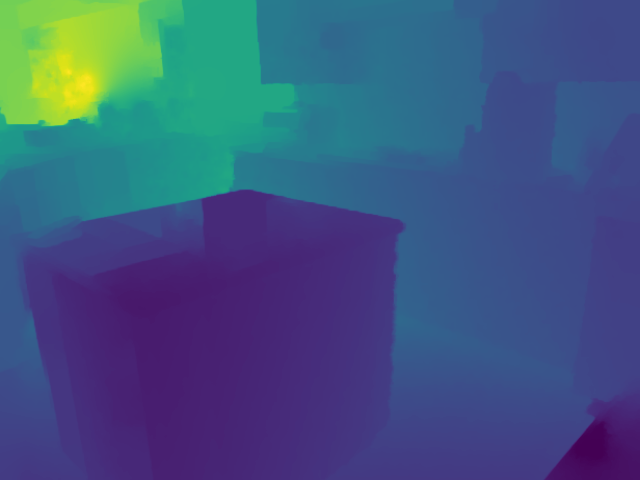} \\
			(a) & (b) \\
			\icg{0.9}{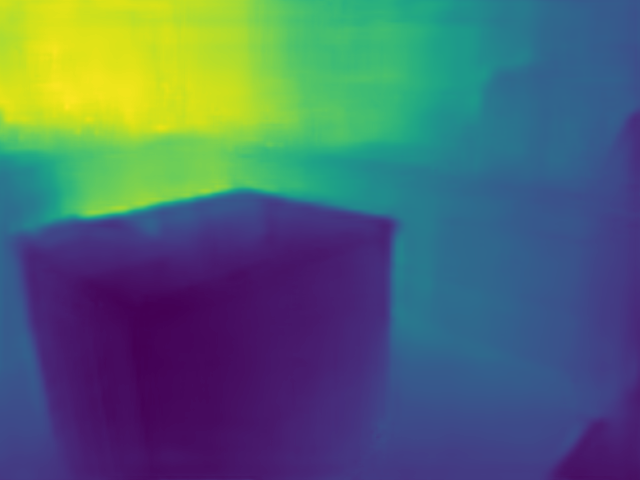} & 
			\icg{0.9}{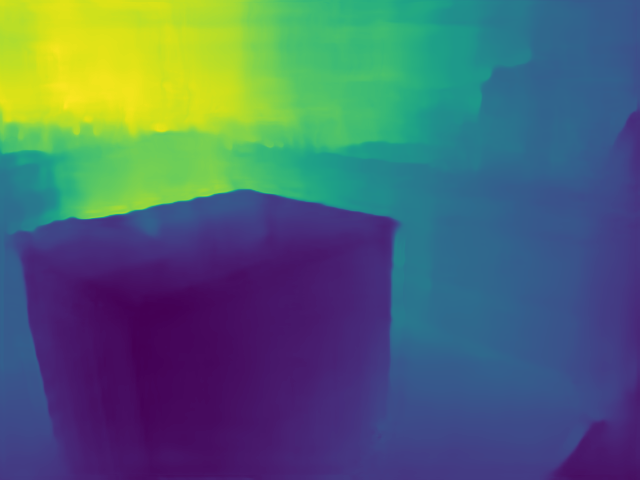} \\
			(c) & (d) \\
		\end{tabular}
	\end{center}
	\label{fig:smooth_edges}
	\caption{(a) Input image overlaid with our occlusion boundary~(OB) manual annotation from NYUv2-OC++ in green, (b) Ground truth depth from NYUv2-Depth, (c) Predicted depth using~\cite{SharpNet2019}, (d) Refined depth using our pixel displacement method.
	}
\end{figure}

\begin{figure}
	\begin{center}
	\begin{tabular}{>{\centering\arraybackslash}m{0.205\linewidth}
					>{\centering\arraybackslash}m{0.205\linewidth}
					>{\centering\arraybackslash}m{0.205\linewidth}
					>{\centering\arraybackslash}m{0.205\linewidth}}
		\includegraphics[width=1\linewidth,height=2cm]{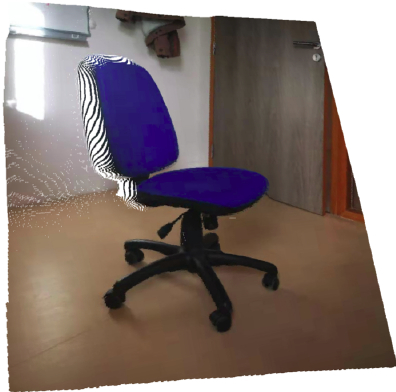} &
		\includegraphics[width=1\linewidth,height=2cm]{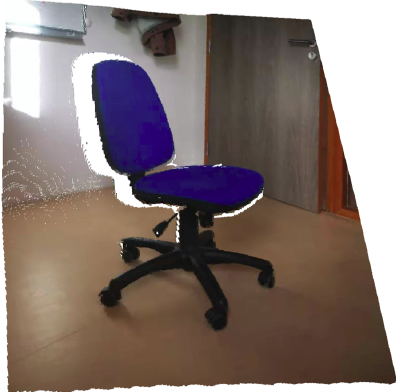} & 
		\includegraphics[width=1\linewidth,height=2cm]{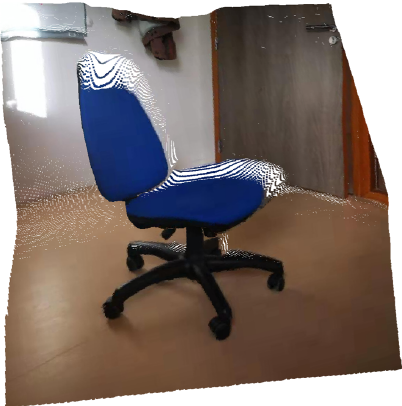} & 
		\includegraphics[width=1\linewidth,height=2cm]{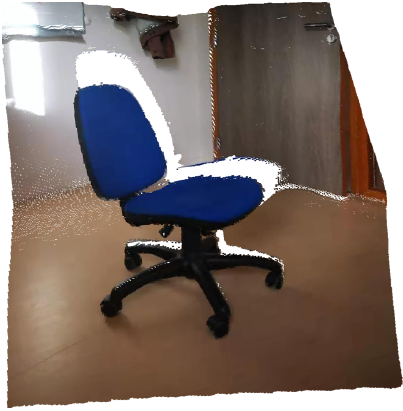} 
		\\
		(a) & (b) & (c) & (d)
	\end{tabular}
	\caption{Application of our depth map refinement to 3D object extraction. (a-b) and (c-d) are two point cloud views of our extracted object. The left column shows point clouds extracted from the initially predicted depth image while the right one shows the result after using our depth refinement method. Our method suppresses long tails around object boundaries, as we achieve sharper occlusion boundaries.}
	\label{fig:chair_application}
	\end{center}
\end{figure}

However,  as demonstrated  in  Fig.~\ref{fig:smooth_edges},  despite the  recent
advances, the  occlusion boundaries  in the predicted  depth maps  remain poorly
reconstructed.  These  occlusion boundaries correspond to  depth discontinuities
happening  along   the  silhouettes  of   objects~\cite{Szeliski97b,  Karsch13}.
Accurate reconstruction of these contours  is thus highly desirable, for example
for handling  partial occlusions between  real and virtual objects  in Augmented
Reality applications or  more  fundamentally,  for  object  understanding,  as  we  show  in
Fig.~\ref{fig:chair_application}.    We   believe   that   this   direction   is
particularly important: Depth prediction generalizes  well to unseen objects and
even  to unseen  object  categories,  and being  able  to  reconstruct well  the
occlusion  boundaries could  be a  promising line  of research  for unsupervised
object discovery.

In this work, we introduce a simple method to overcome smooth occlusion boundaries.  Our method improves their sharpness as well as their localization in the images.  It relies on a differentiable module that takes an initial depth map provided by some depth prediction method, and re-sample it to obtain more accurate occlusion boundaries.  Optionally, it can also take the color image as additional input for guidance information and obtain even better contour localization.  This is done by training a deep network to predict a 2D displacement field, applied to the initial depth map.  This contrasts with previous methods that try to improve depth maps by predicting residual offsets for depth values~\cite{Jeon_2018_ECCV, ZhangPAP-CVPR19}. We show that predicting displacements instead of such a residual helps reaching sharper occluding boundaries.  We believe this is because our module enlarges the family of functions that can be represented by a deep network. As our experiments show, this solution is complementary with all previous solutions as it systematically improves the localization and reconstruction of the occlusion boundaries.

\begin{figure*}[h]
	\icg{1}{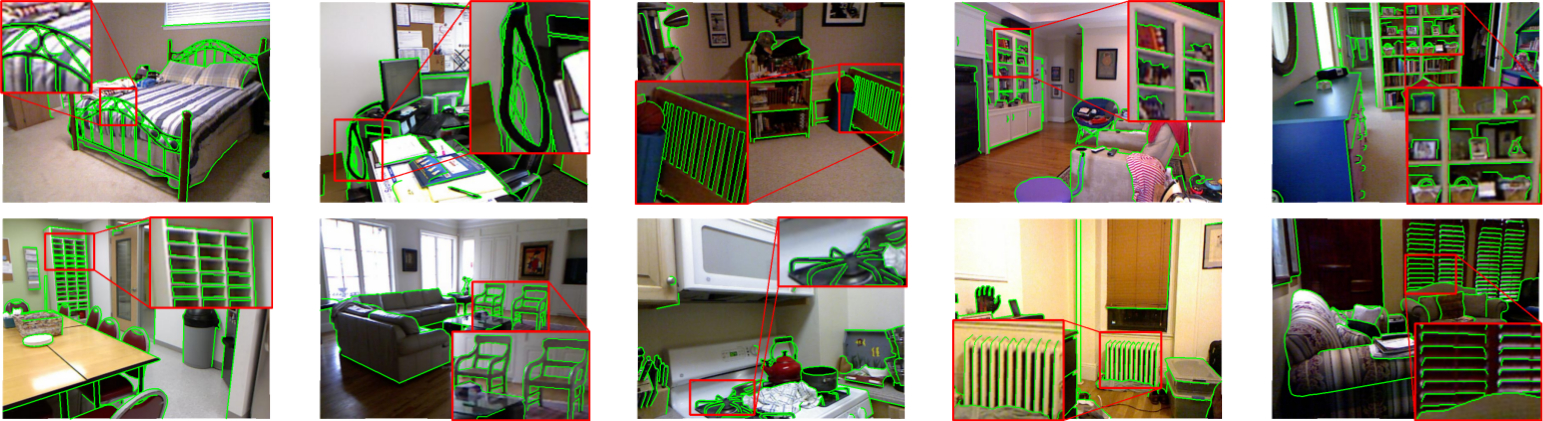}
	\caption{Samples of our NYUv2-OC++ dataset, which extends NYUv2-OC from
          \cite{SharpNet2019}. The selected highlighted regions in red
          rectangles emphasize the high-quality and fine-grained annotations.}
	\label{fig:dataset}
\end{figure*}

In  order  to evaluate  the  occlusion  boundary reconstruction  performance  of
existing  MDE methods  and of  our proposed  method, we  manually annotated  the
occlusion boundaries  in all the images  of the NYUv2 test  set.  Some annotations
are shown in Fig.~\ref{fig:dataset} and  in the supplementary material.  We rely
on  the  metrics  introduced  in  \cite{Koch2018EvaluationOC}  to  evaluate  the
reconstruction  and localization  accuracy  of occlusion boundaries in  predicted
depth maps.  We show that our  method quantitatively improves the performance of
all  state-of-the-art  MDE  methods  in terms  of  localization  accuracy  while
maintaining or improving  the global performance in depth  reconstruction on two
benchmark datasets.

In the rest  of the paper, we  first discuss related work.  We  then present our
approach to sharpen occlusion boundaries in depth maps. Finally, we describe our
experiments and  results to prove  how our method improves  state-of-the-art MDE
methods.

%% file: related.tex

Occlusion boundaries   are  a   notorious  and   important  challenge   in  dense reconstruction from images, for multiple reasons.  For example, it is well known that in  the case of stereo  reconstruction, some pixels in  the neighborhood of occluding  boundaries are  hidden in  one image  but visible  in the  other image, making  the  task  of  pixel  matching difficult.   In  this  context,  numerous solutions     have    already     been     proposed     to    alleviate     this problem~\cite{Fua91a,Intille94,Kanade94,Geiger95}.    We  focus   here  on   recent techniques used in  monocular depth estimation to improve  the reconstruction of occlusion boundaries.

\subsection{Deep Learning for Monocular Depth Estimation~(MDE) and Occlusion Boundaries}

MDE has already been studied thoroughly in the past due to its fundamental and 
practical importance. The surge of deep learning methods and large scale 
datasets has helped it improve continuously, with both supervised and 
self-supervised approaches. To tackle the scale ambiguity discussed in 
MDE discussed in~\cite{Eigen14}, multi-scale deep learning methods have been 
widely used to learn depth priors as they extract global context from an 
image~\cite{Eigen15}. This approach has seen continuous 
improvement as more sophisticated and deeper architectures~\cite{Unet, He16, 
Laina2016DeeperDP} were being developed.

Ordinal regression~\cite{FuCVPR18-DORN} was proposed  as an alternative approach
to direct depth regression in MDE.  This method reached state-of-the-art on both
popular MDE benchmarks  NYUv2-Depth~\cite{Nyuv2} and KITTI~\cite{Geiger2013IJRR}
and  was  later  extended~\cite{Lee2019MonocularDE}.  While  ordinal  regression
helps  reaching  sharper occlusion  boundaries,  those  are unfortunately  often
inaccurately located in the image, as our experiments show.

Another direction to improve the sharpness of object and occlusion boundaries in
MDE is loss  function design. L1-loss and  variants such as the  Huber and BerHu
estimators~\cite{BerHuOwen, BerHuZwald} have now  been widely adopted instead of
L2 since they tend to penalize  discontinuities less.  However, this solution is
not    perfect     and    inaccurate    or    smooth     occlusion    boundaries
remains~\cite{Jiao2018LookDI}.    To  improve   their  reconstruction   quality,
previous work considered depth  gradient matching constraints~\cite{Eigen15} and
depth-to-image   gradient~\cite{Heise2013PMHuber,    Monodepth17}   constraints,
however for the latter, occlusion boundaries  do not always correspond to strong
image gradients, and conversely \textit{e.g.}  for areas with texture gradients.
Constraints   explicitly  based   on   occlusion  boundaries   have  also   been
proposed~\cite{Yang2018UnsupervisedLO,YangLegoCVPR19,SharpNet2019}, leading to a
performance increase in quality of occlusion boundary reconstruction.

Our work is complementary to all of the above approaches, as we show that it 
can improve the localization and reconstruction of occlusion boundaries in all 
state-of-the-art deep learning based methods.

\subsection{Depth Refinement Methods}

In this section we discuss two main  approaches that can be used to refine depth
maps  predictions: We  first  discuss the  formerly  popular Conditional  Random
Fields~(CRFs), then classical filtering methods and their newest versions.

Several previous work  used CRFs post-processing potential to  refine depth maps
predictions. These  works typically define  pixel-wise and pair-wise  loss terms
between  pixels and  their neighbors  using an  intermediate predicted  guidance
signal   such  as   geometric  features~\cite{WangSurgeNIPS16}   or  reliability
maps~\cite{Heo_2018_ECCV}. An initially  predicted depth map is  then refined by
performing inference  with the CRF,  sometimes iteratively or using  cascades of
CRFs~\cite{xu2017multi,xu2018monocular}.   While  most  of  these  methods  help
improving the initial  depth predictions and yield  qualitatively more appealing
results, those methods still  under-perform state-of-the-art non-CRF MDE methods
while being more computationally expensive.

Another option for  depth refinement is to use image  enhancement methods.  Even
though these methods do not  necessarily explicitly target occlusion boundaries,
they can be potential alternative solutions to  ours and we compare with them in
Section~\ref{sec:other_methods}.

Bilateral filtering is a popular and currently state-of-the-art method for image
enhancement,  in particular  as a  denoising method  preserving image  contours.
Although  historically,   it  was  limited   for  post-processing  due   to  its
computational complexity,  recent work have successfully  made bilateral filters
reasonably efficient and fully differentiable~\cite{LiECCV16,wu2017fast}.  These
recent  methods have  been  successful when  applied in  downsampling-upsampling
schemes,  but  have   not  been  used  yet  in  the   context  of  MDE.   Guided
filters~\cite{HePAMI2013_GuidedFilter}  have  been  proposed as  an  alternative
simpler version  of the bilateral filter.  We show in our  experiments that both
guided and bilateral filters sharpen  occlusion boundaries thanks to their usage
of the image for guidance.  They  however sometime produce false depth gradients
artifacts.  The bilateral  solver~\cite{barron2016fast} formulates the bilateral
filtering problem as a  regularized least-squares optimization problem, allowing
fully  differentiable and  much  faster  computation. However,  we  show in  our
experiments that our end-to-end trainable method compares favorably against this
method, both in speed and accuracy.

\subsection{Datasets with Image Contours}

Several datasets of image contours or occlusion boundaries already exist. Popular datasets for edge detection training and evaluation were focused on perceptual~\cite{Martin01, Martin04, Arbelaez11} or object instance~\cite{Everingham10} boundaries. However, those datasets often lack annotation of the occlusion relationship between two regions separated by the boundaries. Other datasets~\cite{Ren2006BSDS_Ownership, HoeimSurfacelayout, Hoeim11, Wang2016PIOD} annotated occlusion relationship between objects, however they do not contain ground truth depth. 

The NYUv2-Depth dataset~\cite{Nyuv2} is a popular MDE benchmark which provides such depth ground truth. Several methods for instance boundary detection have benefited from this depth information~\cite{Gupta13, Dollar13, Ren12, Gupta14, DengECCV18} to improve their performances on object instance boundaries detection. 

The above cited datasets all lack object self-occlusion boundaries and are sometimes inaccurately annotated. Our NYUv2-OC++ dataset provides manual  annotations for the occlusion boundaries on top of NYUv2-Depth for all of its 654 test images. As   discussed  in~\cite{SharpNet2019}, even though it is a  tedious  task,  manual  annotation  is  much  more  reliable  than  automated annotation that could be obtained from depth maps. Fig.~\ref{fig:dataset} illustrates the extensive and accurate coverage of the occlusion  boundaries of our annotations. This dataset enables simultaneous evaluation of depth estimation methods and occlusion boundary reconstruction as the 100 images iBims dataset~\cite{Koch2018EvaluationOC}, but is larger and has been widely used for MDE evaluation.

%% file: method.tex

\input{notations}

In  this section,  we  introduce  our occlusion  boundary  refinement method  as
follows.  Firstly, we present  our hypothesis on the typical structure
  of predicted  depth maps  around occlusion  boundaries and  derive a  model to
  transform  this  structure  into  the  expected one.    We  then  prove  our
hypothesis using  an hand-crafted method.   Based on  this model, we  propose an
end-to-end trainable module which can resample pixels of an input depth image to
restore its sharp occlusion boundaries.

\subsection{Prior Hypothesis}
\label{ssec:model}

Occlusion boundaries  correspond to  regions in the  image where  depth exhibits
large and sharp variations, while the  other regions tend to vary much smoother.
Due  to the  small proportion  of  such sharp  regions, neural  networks tend  to predict over-smoothed depths in the vicinity of occlusion boundaries.

We show in this work that sharp and accurately located boundaries can be recovered
by resampling pixels in the depth map.  This resampling can be formalized as:
\begin{equation}
\forall \bp \in \Omega, \quad D(\bp)\leftarrow D(\bp + \bdeltap(\bp))
\> ,
\label{eq:displacement_update}
\end{equation}
where $D$ is  a depth map, $\bp$  denotes an image location  in domain $\Omega$,
and $\bdeltap(\bp)$ a  2D displacement that depends on  $\bp$.  This formulation
allows  the  depth  values on  the  two  sides  of  occlusion boundaries  to  be
``stitched'' together  and replace  the over-smoothed  depth values.  While this
method  shares some  insights  with Deformable  CNNs~\cite{DaiICCV17} where  the
convolutional kernels are  deformed to adapt to different shapes,  our method is
fundamentally different as we displace depth  values using a predicted 2D vector
for each image location.

Another option to improve depth values  would be to predict an additive residual
depth, which can be formalized as, for comparison:
\begin{equation}
\forall \bp \in \Omega, \quad D(\bp)\leftarrow D(\bp) +
\Delta\!D(\bp) \> .
\label{eq:residual_update}
\end{equation}
We argue that  updating the predicted depth $\hbD$ using  predicted pixel shifts
to  recover  sharp  occlusion  boundaries  in depth  images  works  better  than
predicting  the residual  depth.  We  validate this  claim with  the experiments
presented  below  on  toy  problems,  and   on  real  predicted  depth  maps  in
Section~\ref{sec:experiments}.

\subsection{Testing the Optimal Displacements}
\label{ssec:golden}

To first validate  our assumption that a displacement  field $\bdeltap(\bp)$ can
improve the  reconstructions of  occlusion boundaries,  we estimate  the optimal
displacements using ground truth depth for several predicted depth maps as:
\begin{equation}
\begin{aligned}
\forall \bp \in \Omega,
\bdeltap^* = \argmin\limits_{\bdeltap : \>\> \bp + \bdeltap \>\> \in \>
	\mathcal{N}(\bp)} (D(\bp) - \widehat{D}(\bp + {\bdeltap}) )^2 \> .
\end{aligned}
\label{eq:gold_standard}
\end{equation}

\begin{figure}
	\begin{center}
	\begin{tabular}{>{\centering\arraybackslash}m{0.27\linewidth}
			>{\centering\arraybackslash}m{0.27\linewidth}
			>{\centering\arraybackslash}m{0.27\linewidth}}
		\icg{1.12}{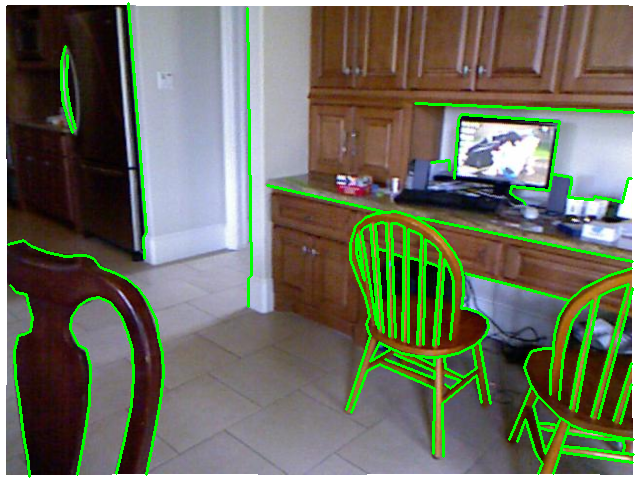} & 
		\icg{1.12}{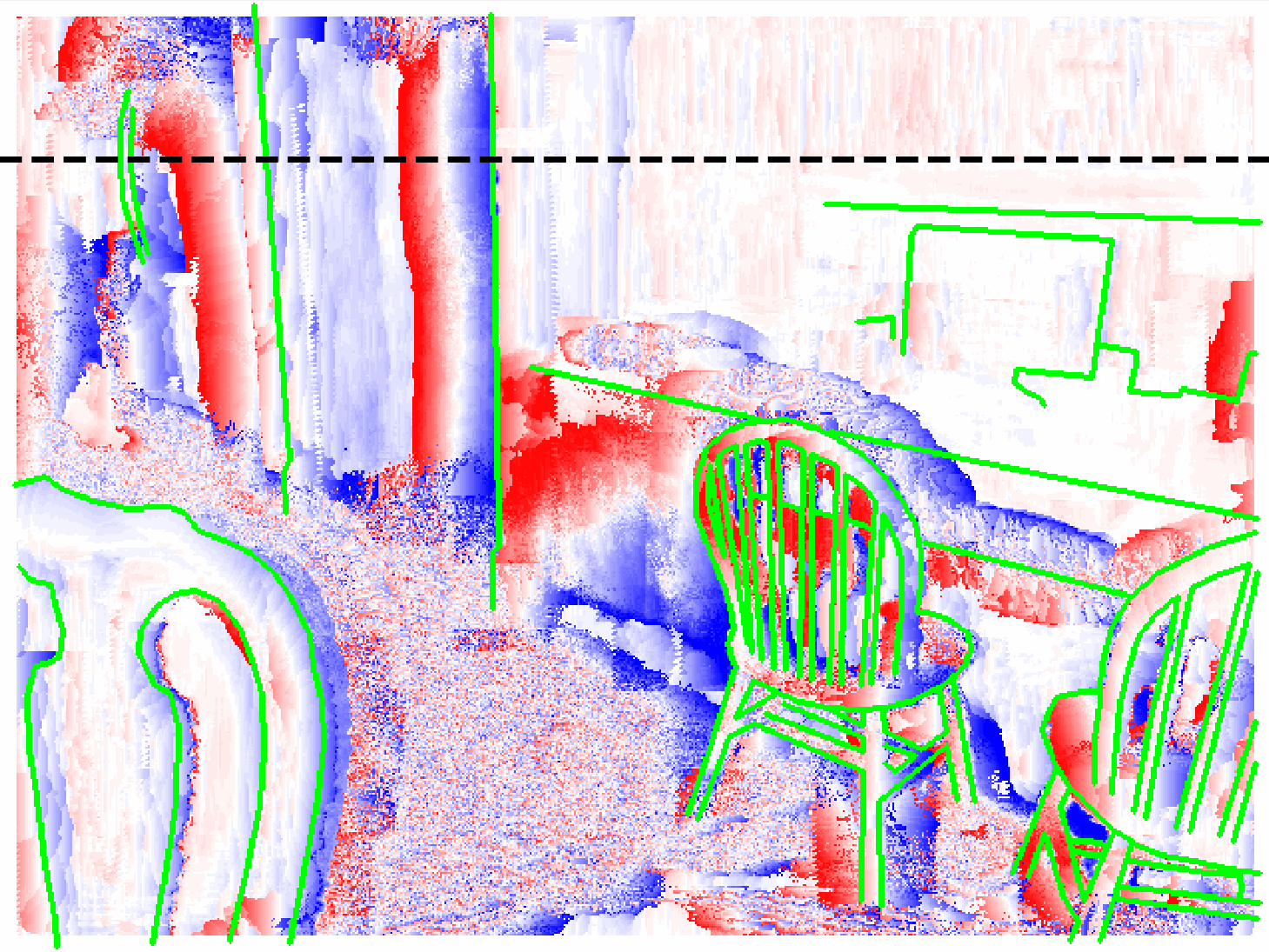} &
		\icg{1.12}{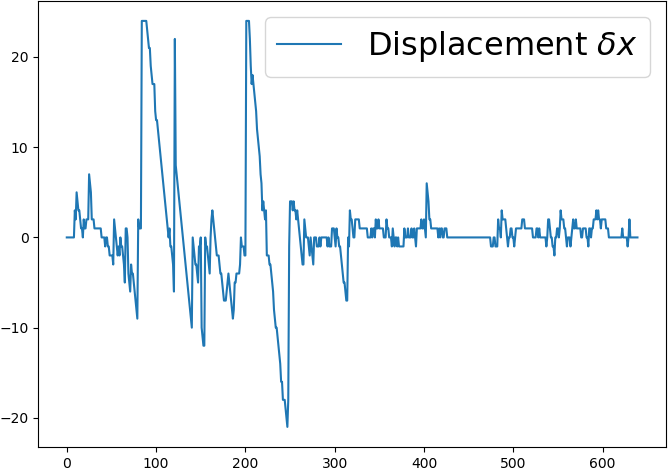} \\
		(a) & (b) & (c) \\
		\icg{1.12}{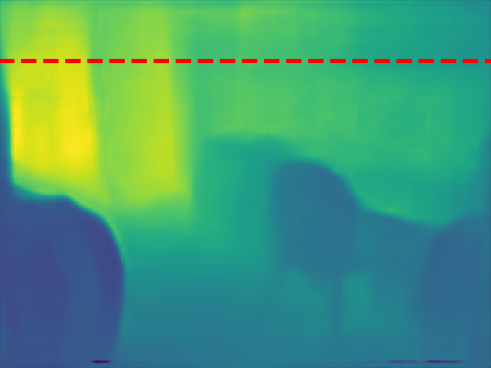} & 
		\icg{1.12}{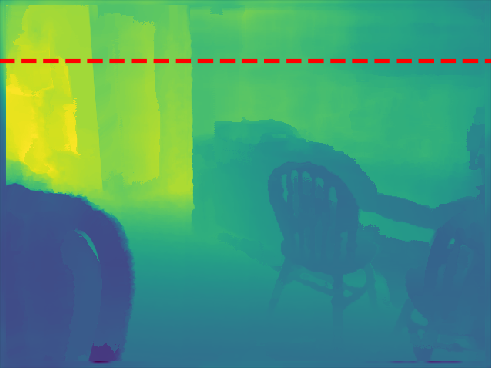} &
		\icg{1.12}{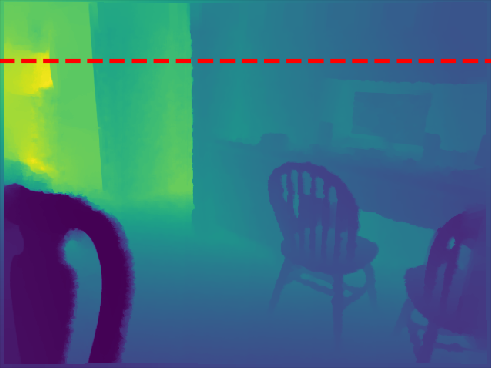} \\	
		(d) & (e) & (f) \\ 	
	\end{tabular}
	\end{center}
	\caption{Refinement  results using  the  gold standard  method
          described   in  Section~\ref{ssec:golden} to   recover  the   optimal
          displacement field (best  seen in color).  (a) is the  input RGB image
          with  superimposed   NYUv2-OC++  annotation  in  green   and  (d)  its
          associated   Ground    Truth   depth.     (e)   is    the   prediction
          using~\cite{Laina2016DeeperDP}  with  pixel  displacements  $\bdeltap$
          from Eq.~\eqref{eq:gold_standard} and (f) the refined prediction.  (b)
          is the  horizontal component of the  displacement field $\bdeltap^{*}$
          obtained by Eq.~\eqref{eq:gold_standard}.  Red and blue color indicate
          positive  and negative  values  respectively.  (c)  is the  horizontal
          component $\delta  x$ of  displacement field $\bdeltap^{*}$  along the
          dashed red line drawn in (b,c,d,e).  }
	\label{fig:gold_standard}
\end{figure}

In  words, solving  this problem  is equivalent  to finding  for each  pixel the
optimal displacement $\bdeltap^*$  that reconstructs the ground  truth depth map
$D$    from    a    predicted     depth    map    $\widehat{D}$.     We    solve
Eq.~\eqref{eq:gold_standard} by  performing for  all pixels $\bp$  an exhaustive
search   of  $\bdeltap$   within  a   neighborhood~$\mathcal{N}(\bp)$  of   size
$50\times50$.   Qualitative results  are shown  in Fig.~\ref{fig:gold_standard}.
The depth  map obtained by applying  this optimal displacement field  is clearly
much better.

In  practice, we  will have  to  predict the  displacement field  to apply  this
idea. This  is done by training  a deep network,  which is detailed in  the next
subsection.  We then check on a  toy problem that this yields better performance
than predicting residual depth with a deep network of similar capacity.

\subsection{Method Overview}

\begin{figure}
	\begin{center}
		\icg{1}{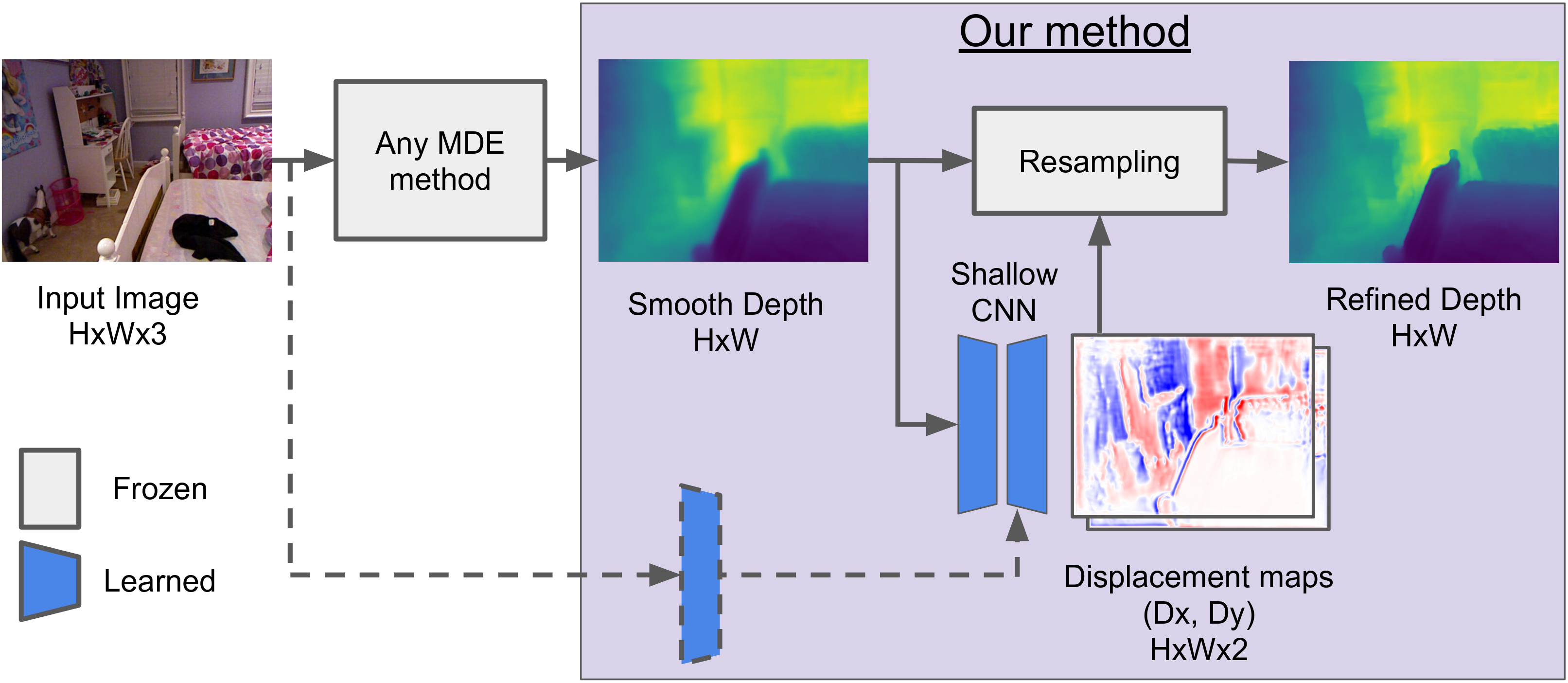}				
		\caption{Our proposed pipeline for depth edge sharpening. The dashed 
			lines define the optional guidance with RGB features for our shallow 
			network. 
		}	
		\label{fig:pipeline}
	\end{center}	
\end{figure}

Based on our model, we propose to learn the displacements of pixels in predicted
depth    images    using    CNNs.     Our    approach    is    illustrated    in
Fig.~\ref{fig:pipeline}:  Given  a  predicted  depth  image  $\widehat{D}$,  our
network predicts a displacement field  $\bdeltap$ to resample the image locations
in  depth  map  $\widehat{D}$ according  to  Eq.~\eqref{eq:displacement_update}.
This    approach    can    be    implemented   as    a    Spatial    Transformer
Network~\cite{Jaderberg15}.

Image guidance  can be helpful  to improve  the occlusion boundary  precision in
refined depth maps and can also help  to discover edges that were not visible in
the initial predicted depth map $\widehat{D}$.  However, it should be noted that
our network can still work even without image guidance.

\subsection{Training Our Model on Toy Problems}
\label{ssec:toy_problem}

In   order  to   verify   that  displacement   fields   $\bdeltap$  presented   in
Section~\ref{ssec:model} can  be learned, we first  define a toy problem  in 1D.

\begin{figure}[t]
\begin{center}
	\icg{0.8}{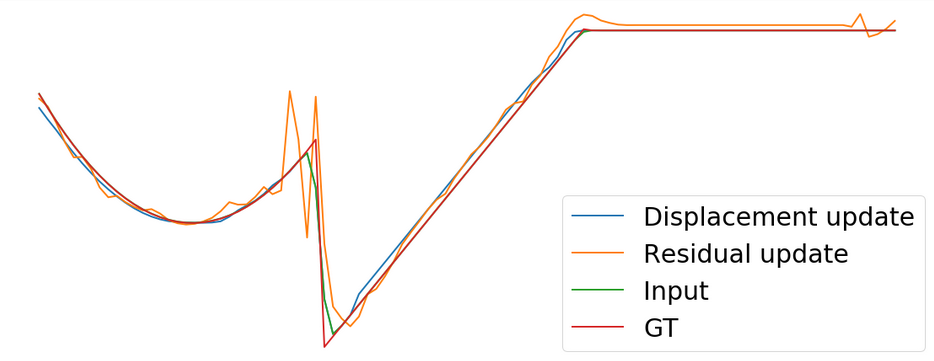}
	\caption{Comparison between displacement and residual update
          learning. Both residual and displacement learning can predict sharper
          edges, however residual updates often produce artifacts along the
          edges while displacements update does not.}
	\label{fig:artifacts_1D}
\end{center}
\end{figure}

In  this toy  problem, as  shown  in Fig.~\ref{fig:artifacts_1D},  we model  the
signals  $D$ to  be recovered  as piecewise  continuous functions,  generated as
sequences  of basic  functions such  as  step, affine  and quadratic  functions.
These samples exhibit strong discontinuities  at junctions and smooth variations
everywhere  else, which  is a  property  similar to  real depth  maps.  We  then
convolve the $D$ signals with  random-size (blurring) Gaussian kernels to obtain
smooth  versions  $\widehat{D}$.  This  gives  us  a  training  set  $\calT$  of
$(\widehat{D}, D)$ pairs.

We use $\calT$ to train a network $f(.;\Theta_f)$ of parameters to predict a
displacement field:
\begin{equation}
\min_{\Theta_f} \sum_{(\widehat{D}, D)\in\calT} \sum_{{\bp}} L\left(D(\bp) -
\widehat{D}\left(\bp+f(\widehat{D};\Theta_f)(\bp)\right)\right) \> .
\end{equation}
and a network $g(.;\Theta_g)$ of parameters to predict a
residual depth map:
\begin{equation}
\min_{\Theta_g} \sum_{(\widehat{D}, D)\in\calT} \sum_{{\bp}} L\left(D(\bp) -
\widehat{D}(\bp) + g(\widehat{D};\Theta_g)(\bp)\right) \> ,
\end{equation}
where $L(.)$ is some loss. In our experiments, we evaluate the $l_1$, $l_2$, and
Huber losses.

As shown in Fig.~\ref{fig:artifacts_1D}, we found that predicting a residual
update produces severe artifacts around  edges such as overshooting effects.  We
argue that these  issues arise because the  residual CNN $g$ is  also trained on
regions where the  values of $\widehat{D}$ and $D$ are  different even away from
edges, thus encouraging network $g$ to also correct these regions.  By contrast,
our method does  not create such overshooting  effects as it does  not alter the
local range of values around edges.

It   is  worth   noticing  that   even  when   we  allow   $\widehat{D}$ and
$D$  to have  slightly different
values in non-edge  areas -which simulates residual error  between predicted and
ground truth depth-, our method still  converges to a good solution, compared to
the residual CNN.

We extend our approach validation from 1D to 2D, where the 1D signal is replaced
by 2D  images with different  polygons of different  values.  We apply  the same
operation to smooth the images and then  use our network to recover the original
sharp  images from  the smooth  ones.  We observed  similar results  in 2D:  The
residual CNN always generates artifacts.  Some results of our 2D toy problem can
be found in supplementary material.

\subsection{Learning to Sharpen Depth Predictions}

To  learn to  produce sharper  depth predictions  using displacement  fields, we
first trained our  method in a similar  fashion to the toy  problem described in
Section~\ref{ssec:toy_problem}.   While this  already  improves  the quality  of
occlusion boundaries of  all depth map predictions, we show  that we can further
improve quantitative results by training our  method using predictions of an MDE
algorithm on the NYUv2-Depth dataset as input and the corresponding ground truth
depth as target output.   We argue that this way, the  network learns to correct
more  complex  distortions of  the  ground  truth  than Gaussian  blurring.   In
practice, we used the predictions of~\cite{SharpNet2019} to demonstrate this, as
it is state-of-the-art  on occlusion boundary sharpness.  We show  that this not
only improves the quality of depth maps predicted by~\cite{SharpNet2019} but all
other available MDE  algorithms.  Training our network using  the predictions of
other algorithms  on the NYUv2 would  further improve their result,  however not
all  of  them   provide  their  code  or  their  predictions   on  the  official
\textit{training set}.   Finally, our  method is  fully differentiable.
While we do not do it in this paper, it is also possible to train an MDE jointly
with our method, which should yield even better results.

\begin{figure*}
	\begin{tabular}{>{\centering\arraybackslash}m{0.14\textwidth}
			>{\centering\arraybackslash}m{0.14\textwidth}
			>{\centering\arraybackslash}m{0.14\textwidth}
			>{\centering\arraybackslash}m{0.14\textwidth}
			>{\centering\arraybackslash}m{0.14\textwidth}
			>{\centering\arraybackslash}m{0.14\textwidth}}
		\multicolumn{6}{c}{\hspace{-0em}\icg{0.97}{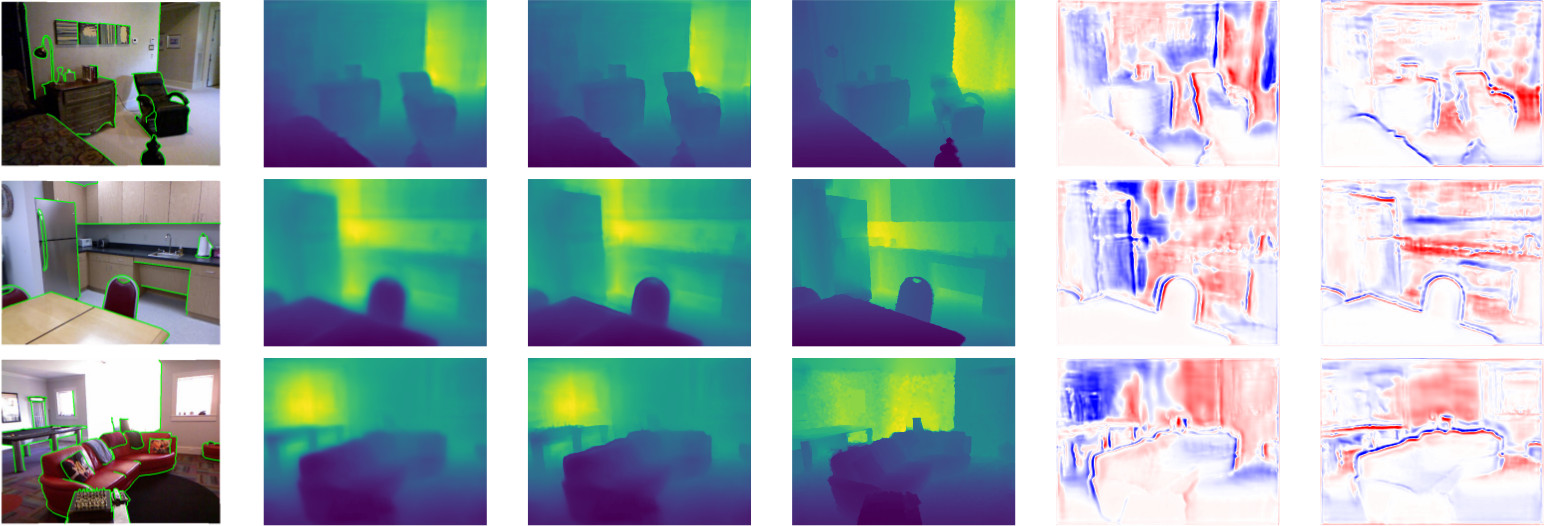}} \\
		$\>\>\>\>$(a) & $\>\>\>\>$(b) & $\>\>\>\>$(c) & $\>\>\>\>$(d) & $\>\>\>\>$(e) & $\>\>\>\>$(f)
	\end{tabular}
	\caption{Refinement results using our method (best seen in color).  From
          left  to right:  (a) input  RGB  image with  NYUv2-OC++ annotation  in
          green, (b) SharpNet~\cite{SharpNet2019}  depth prediction, (c) Refined
          prediction, (d)  Ground truth depth,  (e) Horizontal and  (f) Vertical
          components of the displacement  field. Displacement fields are clipped
          between $\pm$15 pixels.  Although SharpNet is used as  an example here
          because  it  is  currently   state-of-the-art  on  occlusion  boundary
          accuracy, similar  results can  be observed when  refining predictions
          from other methods.}
	\label{fig:results}
\end{figure*}

%% file: notations.tex
\newcommand{\bhC}{\widehat{\boldsymbol{C}}}
\newcommand{\hCi}{\widehat{C}_i}
\newcommand{\bhD}{\widehat{\boldsymbol{D}}}
\newcommand{\hDi}{\widehat{D}_i}
\newcommand{\bhN}{\widehat{\boldsymbol{N}}}
\newcommand{\bhNi}{\widehat{\boldsymbol{N}}_i}

\newcommand{\bC}{\boldsymbol{C}}
\newcommand{\bD}{\boldsymbol{D}}
\newcommand{\bI}{\boldsymbol{I}}
\newcommand{\bN}{\boldsymbol{N}}
\newcommand{\bM}{\boldsymbol{M}}
\newcommand{\bX}{\boldsymbol{X}}
\newcommand{\bn}{\boldsymbol{n}}
\newcommand{\bp}{\boldsymbol{p}}
\newcommand{\bu}{\boldsymbol{u}}
\newcommand{\bdeltap}{\boldsymbol{\delta p}}
\newcommand{\hbD}{\widehat{\bD}}

\newcommand{\calT}{\mathcal{T}}
\newcommand{\calL}{\mathcal{L}}

\newcommand{\normals}{\text{n}}
\newcommand{\depth}{\text{d}}
\newcommand{\boundaries}{\text{c}}
\newcommand{\superv}{\text{superv}}
\newcommand{\dc}{\text{dc}}
\newcommand{\dn}{\text{dn}}
\newcommand{\BerHu}{\text{BerHu}}
\newcommand{\Huber}{\text{Huber}}
\newcommand{\AL}{\text{AL}}

\newcommand{\IR}{\mathbb{R}}

%% file: experiments.tex

\label{sec:experiments}

In this section, we first detail our implementation, describe the metrics we use
to evaluate the  reconstruction of the occlusion boundaries and  the accuracy of
the depth  prediction, and then  present the results  of the evaluations  of our
method on the outputs from different MDE methods.

\subsection{Implementation Details}
\label{implementation details}

We   implemented  our   network  using   the  Pytorch
framework~\cite{paszke2017automatic}.  Our network is  a light-weight encoder-decoder network with
  skip connections,  which has one encoder for depth, an optional one for
  guidance with the RGB image, and a shared decoder.  Details on each component can be found in
the supplementary  material. We  trained  our network  on the  output  of a  MDE
method~\cite{SharpNet2019},  using Adam  optimization with  an initial  learning
rate  of 5e-4  and weight  decay of  1e-6, and  the \textit{poly}  learning rate
policy~\cite{zhao2017pspnet}, during 32k iterations on NYUv2~\cite{Nyuv2}.  This
dataset contains  1449 pairs RGB  and depth images,  split into 795  samples for
training and 654 for  testing.  Batch size was set to 1.   The input images were
resized  with  scales  $[0.75, 1,  1,5,  2]$  and  then  cropped and  padded  to
$320\times320$.  We tried  to learn on the \emph{raw} depth  maps from the NYUv2
dataset, without success.  This is most likely due to the fact that missing data
occurs mostly around  the occlusion boundaries.  Dense depth  maps are therefore
needed which is  why we use~\cite{Levin2004colorization} to  inpaint the missing
data in the raw depth maps.

\subsection{Evaluation metrics}
\label{metrics}

\subsubsection{Evaluation of monocular depth prediction}
As for previous work~\cite{Eigen14, Eigen15, Laina2016DeeperDP}, we evaluate the monocular depth predictions using the following metrics: Root Mean Squared linear Error~($R_{lin}$), mean absolute relative error~(rel), mean log$_{10}$ error~(log$_{10}$), Root Mean Squared log Error~($R_{log}$), and the accuracy under threshold~($\sigma_{i}$ $<$ 1.25$^{i}$, $i = 1,2,3$).

\subsubsection{Evaluation of occlusion boundary accuracy}
\begin{figure}[t]
	\begin{center}
		\begin{tabular}{>{\centering\arraybackslash}m{0.49\linewidth}
				>{\centering\arraybackslash}m{0.49\linewidth}
			}
			\icg{0.7}{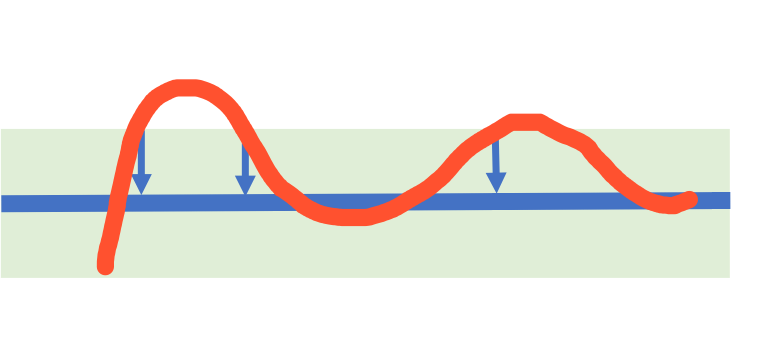} &		 
			\icg{0.7}{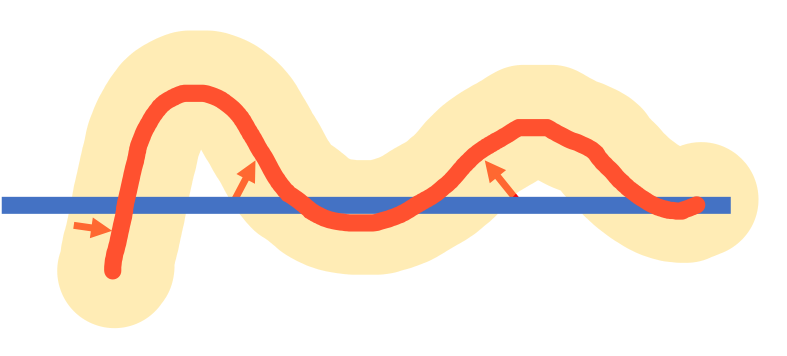} \\
			(a) & (b) \\
		\end{tabular}
	\end{center}	
	\caption{Occlusion boundary evaluation metrics based on the
          Chamfer distance, as introduced in
          \cite{Koch2018EvaluationOC}. The blue lines represent the
          ground truth boundaries, the red curve the predicted
          boundary. Only the boundaries in the green area are taken
          into account during the evaluation of accuracy (a), and only
          the yellow area are taken into account during the evaluation
          of completeness (b). (a) The accuracy is evaluated from the
          distances of points on the predicted boundaries to the
          ground truth boundaries. (b) The completeness is evaluated
          from the distances of points on the ground truth boundaries to the
          predicted boundaries. }
	\label{fig:chamfer_distance}
\end{figure}

Following  the  work  of Koch  \textit{et  al.}~\cite{Koch2018EvaluationOC},  we
evaluate  the accuracy  of occlusion  boundaries~(OB) using  the proposed  depth
boundary  errors,  which evaluate  the  accuracy  $\epsilon_{a}$ and  completion
$\epsilon_{c}$  of  predicted occlusion  boundaries.  The  boundaries are  first
extracted using a Canny  edge detector~\cite{Canny86} with predefined thresholds
on    a    normalized    predicted    depth   image.     As    illustrated    in
Fig.~\ref{fig:chamfer_distance}, $\epsilon_{a}$ is taken  as the average Chamfer
distance  in pixels~\cite{Fan_2017_CVPR}  from the  predicted boundaries  to the
ground  truth  boundaries; $\epsilon_{com}$  is  taken  as the  average  Chamfer
distance from ground truth boundaries to the predicted boundaries.

\subsection{Evaluation on the NYUv2 Dataset}

We evaluate our method by refining the predictions of different state-of-the-art methods~\cite{Eigen14, Laina2016DeeperDP, FuCVPR18-DORN, SharpNet2019, Jiao2018LookDI, Yin2019enforcing}. Our network is trained using the 795 labeled NYUv2 depth images of training dataset with corresponding RGB images as guidance. 

To enable a fair comparison, we evaluate only pixels inside the crop defined in~\cite{Eigen14} for all methods. Table~\ref{tab:oc} shows the evaluation of refined predictions of different methods using our network. With the help of our proposed network, the occlusion boundary accuracy of all methods can be largely improved, without degrading the global depth estimation accuracy. We also show qualitative results of our refinement method in Fig.~\ref{fig:results}.

\newcommand{\mrt}{1pt}

\begin{table*}
\centering
\ra{0.8}
\renewcommand\tabcolsep{8pt}
\begin{tabular}{@{}lcccccccccccc@{}}
	\toprule[\mrt]
	~ & ~ & \multicolumn{4}{c}{Depth error~($\downarrow$)} && \multicolumn{3}{c}{Depth accuracy~($\uparrow$)} &&
	\multicolumn{2}{c}{OBs~($\downarrow$)} \\		
	\cmidrule(l{7pt}r{10pt}){3-6} \cmidrule(l{7pt}r{10pt}){8-10} \cmidrule(l{7pt}){12-13}
Method & Refine & rel & log$_{10}$ & $R_{lin}$ & $R_{log}$ && $\sigma_{1}$ & $\sigma_{2}$ & $\sigma_{3}$ && $\epsilon_{a}$ & $\epsilon_{c}$ \\
\midrule[\mrt]
\multirow{2}*{Eigen \textit{et al.}~\cite{Eigen14}}
& - & 0.234 & 0.095 & 0.760 & 0.265 && 0.612 & 0.886 & 0.971 && 9.936 & 9.997 \\
& \checkmark & 0.232 & 0.094 & 0.758 & 0.263 && 0.615 & 0.889 & 0.971 && 2.168 & 8.173 \\
\midrule
\multirow{2}*{Laina \textit{et al.}~\cite{Laina2016DeeperDP}} 
& - & 0.142 & 0.059 & 0.510 & 0.181 && 0.818 & 0.955 & 0.988 && 4.702 & 8.982 \\
& \checkmark & 0.140 & 0.059 & 0.509 & 0.180 && 0.819 & 0.956 & 0.989 && 2.372 & 7.041 \\
\midrule
\multirow{2}*{Fu \textit{et al.}~\cite{FuCVPR18-DORN}} 
& - & 0.131 & 0.053 & 0.493 & 0.174 && 0.848 & 0.956 & 0.984 && 3.872 & 8.117 \\
& \checkmark & 0.136 & 0.054 & 0.502 & 0.178 && 0.844 & 0.954 & 0.983 && 3.001 & 7.242 \\
\midrule
Ramamonjisoa 
& - & 0.116 & 0.053 & 0.448 & 0.163 && 0.853 & 0.970 & 0.993 && 3.041 & 8.692 \\
and Lepetit~\cite{SharpNet2019} & \checkmark & 0.117 & 0.054 & 0.457 & 0.165 && 0.848 & 0.970 & 0.993 && 1.838 & 6.730 \\
\midrule
\multirow{2}*{Jiao \textit{et al.}~\cite{Jiao2018LookDI}} 
& - & 0.093 & 0.043 & 0.356 & 0.134 && 0.908 & \textbf{0.981} &\textbf{ 0.995} && 8.730 & 9.864 \\
& \checkmark & \textbf{0.092} & \textbf{0.042} & \textbf{0.352} & \textbf{0.132} && \textbf{0.910} & \textbf{0.981} & \textbf{0.995} && 2.410 & 8.230 \\
\midrule
\multirow{2}*{Yin \textit{et al.}~\cite{Yin2019enforcing}} 
& - & 0.112 & 0.047 & 0.417 & 0.144 && 0.880 & 0.975 & 0.994 && 1.854 & 7.188 \\
& \checkmark & 0.112 & 0.047 & 0.419 & 0.144 && 0.879 & 0.975 & 0.994 && \textbf{1.762} & \textbf{6.307} \\
\bottomrule[\mrt]
\end{tabular}
\vspace{2pt}
\caption{Evaluation  of our  method on  the output  of several  state-of-the-art
  methods on NYUv2.  Our method significantly improves  the occlusion boundaries
  metrics  $\epsilon_{a}$  and  $\epsilon_{c}$ without  degrading  the  other
  metrics related to the overall depth accuracy. These results were computed
  using available depth maps predictions (apart from Jiao~\textit{et
    al.}~\cite{Jiao2018LookDI} who sent us their predictions) within the image region proposed in~\cite{Eigen14}. ($\downarrow$: Lower is better; $\uparrow$:
  Higher is better).}
\label{tab:oc}
\end{table*}

\subsection{Evaluation on the iBims Dataset}
We applied  our method trained on NYUv2 dataset to refine the predictions from various methods~\cite{Eigen14, Eigen15, Laina2016DeeperDP, Depth2015CVPR, LiICCV17, Liu2018planenet, SharpNet2019} on the iBims dataset~\cite{Koch2018EvaluationOC}. Table~\ref{tab:ibims} shows that our network significantly improves the accuracy and completeness metrics for the occluding boundaries of all predictions on this dataset as well. 

\begin{table*}
	\centering
	\ra{0.8}
	\renewcommand\tabcolsep{10pt}
	\begin{tabular}{@{}lccccccccccc@{}}
		\toprule[\mrt]
		~ & ~ & \multicolumn{3}{c}{Depth error~($\downarrow$)} && \multicolumn{3}{c}{Depth accuracy~($\uparrow$)} &&
		\multicolumn{2}{c}{OB~($\downarrow$)} \\		
		\cmidrule(l{9pt}r{10pt}){3-5} \cmidrule(l{9pt}r{10pt}){7-9} \cmidrule(l{10pt}){11-12}
		Method & Refine & rel & log$_{10}$ & $R_{lin}$ && $\sigma_{1}$ & $\sigma_{2}$ & $\sigma_{3}$ && $\epsilon_{a}$ & $\epsilon_{c}$ \\
		\midrule[\mrt]
		\multirow{2}*{Eigen \textit{et al.}~\cite{Eigen14}}
		& - & 0.32 & 0.17 & 1.55 && 0.36 & 0.65 & 0.84 && 9.97 & 9.99\\
		& \checkmark & 0.32 & 0.17 & 1.54 && 0.37 & 0.66 & 0.85 && 4.83 & 8.78\\
		\midrule
		Eigen and Fergus 
		& - & 0.30 & 0.15 & 1.38 && 0.40 & 0.73 & 0.88 && 4.66 & 8.68\\
		(AlexNet)~\cite{Eigen15} & \checkmark & 0.30 & 0.15 & 1.37 && 0.41 & 0.73 & 0.88 && 4.10 & 7.91\\ 
		\midrule
		Eigen and Fergus
		& - & 0.25 & 0.13 & 1.26 && 0.47 & 0.78 & 0.93 && 4.05 & 8.01\\
		(VGG)~\cite{Eigen15} & \checkmark & 0.25 & 0.13 & 1.25 && 0.48 & 0.78 & 0.93 && 3.95 & 7.57\\
		\midrule
		\multirow{2}*{Laina \textit{et al.}~\cite{Laina2016DeeperDP}} 
		& - & 0.26 & 0.13 & 1.20 && 0.50 & 0.78 & 0.91 && 6.19 & 9.17\\
		& \checkmark & 0.25 & 0.13 & 1.18 && 0.51 & 0.79 & 0.91 && 3.32 & 7.15\\
		\midrule
		\multirow{2}*{Liu \textit{et al.}~\cite{Depth2015CVPR}} 
		& - & 0.30 & 0.13 & 1.26 && 0.48 & 0.78 & 0.91 && 2.42 & 7.11\\
		& \checkmark & 0.30 & 0.13 & 1.26 && 0.48 & 0.77 & 0.91 && 2.36 & 7.00\\
		\midrule
		\multirow{2}*{Li \textit{et al.}~\cite{LiICCV17}}
		& - & \textbf{0.22} & \textbf{0.11} & 1.09 && 0.58 & \textbf{0.85} & \textbf{0.94} && 3.90 & 8.17\\
		& \checkmark & \textbf{0.22} & \textbf{0.11} & 1.10 && 0.58 & 0.84 & \textbf{0.94} && 3.43 & 7.19\\
		\midrule
		\multirow{2}*{Liu \textit{et al.}~\cite{Liu2018planenet}} 
		& - & 0.29 & 0.17 & 1.45 && 0.41 & 0.70 & 0.86 && 4.84 & 8.86\\
		& \checkmark & 0.29 & 0.17 & 1.47 && 0.40 & 0.69 & 0.86 && 2.78 & 7.65\\
		\midrule
		Ramamonjisoa
		& - & 0.27 & \textbf{0.11} & \textbf{1.08} && \textbf{0.59} & 0.83 & 0.93 && 3.69 & 7.82\\
		and Lepetit~\cite{SharpNet2019} & \checkmark & 0.27 & \textbf{0.11} & \textbf{1.08} && \textbf{0.59} & 0.83 & 0.93 && \textbf{2.13} & \textbf{6.33}\\
		\bottomrule[\mrt]
	\end{tabular}
	\vspace{2pt}
	\caption{Evaluation  of our  method on  the output  of several
          state-of-the-art methods on iBims.}
	\label{tab:ibims}
\end{table*}

\subsection{Comparison with Other Methods}
\label{sec:other_methods}
To demonstrate the efficiency of our proposed method, we compare our method with
existing  filtering methods~\cite{tomasi1998bilateral,  HePAMI2013_GuidedFilter,
  barron2016fast, wu2017fast}. We use the prediction of \cite{Eigen14} as input,
and compare  the accuracy of depth  estimation and occlusion boundaries  of each
method.  Note that for the filters  with hyper-parameters, we tested each filter
with a series  of hyper-parameters and select the best  refined results. For the
Fast  Bilateral Solver~(FBS)~\cite{barron2016fast}  and the  Deep Guided  Filter
(GF)~\cite{wu2017fast},  we  use  their   default  settings  from  the  official
implementation.  We  keep the same network  with and without Deep  GF, and train
both times with the same learning rate and data augmentation.

As shown  in Table~\ref{tab:methods  comparison}, our  method achieves  the best
accuracy for occlusion  boundaries.  Finally, we compare our  method against the
additive residual prediction method  discussed in~\ref{ssec:model}.  We keep the
same U-Net architecture, but replace the displacement operation with an addition
as described in  Eq.~\eqref{eq:residual_update}, and show that  we obtain better
results. We  argue that the performance  of the deep guided  filter and additive
residual  are  lower   due  to  generated  artifacts  which   are  discussed  in
Section~\ref{ssec:toy_problem}.   In   Table~\ref{tab:fps_comparison},  we  also
compare  our   network's  computational   efficiency  against   reference  depth
estimation and refinement methods.

\begin{table}[h]
	\begin{scriptsize}	
		\begin{center}
			\ra{0.7}
			\renewcommand\tabcolsep{2.8pt}
			\begin{tabular}{@{}
					P{0.17\linewidth}
					>{\centering\arraybackslash}m{0.06\linewidth}
					>{\centering\arraybackslash}m{0.06\linewidth}
					>{\centering\arraybackslash}m{0.06\linewidth}
					>{\centering\arraybackslash}m{0.08\linewidth}
					>{\centering\arraybackslash}m{0.06\linewidth}
					>{\centering\arraybackslash}m{0.06\linewidth}
					>{\centering\arraybackslash}m{0.07\linewidth}
					>{\centering\arraybackslash}m{0.07\linewidth}
					>{\centering\arraybackslash}m{0.07\linewidth}@{}
				}
				\toprule[\mrt]
				~ & \multicolumn{4}{c}{Depth error~($\downarrow$)} & \multicolumn{3}{c}{Depth accuracy~($\uparrow$)} & 
				\multicolumn{2}{c}{OB~($\downarrow$)} \\
				\cmidrule(l{3pt}r{4pt}){2-5}
				\cmidrule(l{3pt}r{4pt}){6-8}
				\cmidrule(l{3pt}){9-10}
				Method  & rel & log10 & $R_{lin}$ & $R_{log}$ & $\sigma_{1}$ & $\sigma_{2}$ & $\sigma_{3}$ & $\epsilon_{a}$ & $\epsilon_{c}$ \\
				\midrule[\mrt]
				Baseline~\cite{Eigen14} & 0.234 & 0.095 & 0.766 & 0.265 & 0.610 & 0.886 & \textit{0.971} & 9.926 & 9.993 \\
				\midrule
				Bilateral Filter~\cite{tomasi1998bilateral} & 0.236 & 0.095 & 0.765 & 0.265 & 0.611 & 0.887 & \textit{0.971} & 9.313 & 9.940 \\
				\midrule
				GF~\cite{HePAMI2013_GuidedFilter} & 0.237 & 0.095 & 0.767 & 0.265 & 0.610 & 0.885 & \textit{0.971} & 6.106 & 9.617 \\
				\midrule
				FBS~\cite{barron2016fast} & 0.236 & 0.095 & 0.765 & 0.264 & 0.611 & 0.887 & \textit{0.971} & 5.428 & 9.454 \\
				\midrule
				Deep GF~\cite{wu2017fast} & 0.306 & 0.116 & 0.917 & 0.362 & 0.508 & 0.823 & 0.948 & 4.318 & 9.597 \\
				\midrule
				Residual & 0.286 & 0.132 & 0.928 & 0.752 & 0.508 & 0.807 & 
				0.931 & 5.757 & 9.785\\		
				
				\midrule
				Our Method & \textbf{0.232} & \textbf{0.094} & \textbf{0.757} & \textbf{0.263} & \textbf{0.615} & \textbf{0.889} & \textit{0.971} & \textbf{2.302} & \textbf{8.347} \\
				\bottomrule[\mrt]
			\end{tabular}
		\end{center}
	\end{scriptsize}
	\caption{Comparison with existing methods for image enhancement, adapted to the depth map prediction problems. Our method performs the best for this problem over all the different metrics.}
	\label{tab:methods comparison}
\end{table}

\begin{table}[h]
	\begin{scriptsize}	
		\begin{center}
			\renewcommand\tabcolsep{2.1pt}
			\begin{tabular}{@{}P{0.16\linewidth}
					>{\centering\arraybackslash}m{0.16\linewidth}
					>{\centering\arraybackslash}m{0.16\linewidth}
					>{\centering\arraybackslash}m{0.16\linewidth}
					>{\centering\arraybackslash}m{0.16\linewidth}
					@{}
				}
				\toprule[\mrt]
				Method & SharpNet~\cite{SharpNet2019} & VNL~\cite{Yin2019enforcing} & Deep GF\cite{wu2017fast} & Ours \\
				\midrule
				FPS - GPU & 83.2 $\pm$ 6.0 & 32.2$\pm$ 2.1 & 70.5 $\pm$ 7.5 & \textbf{100.0} $\pm$ 7.3\\
				FPS - CPU & 2.6 $\pm$ 0.0 & * & 4.0 $\pm$ 0.1 & \textbf{5.3} $\pm$ 0.15	\\
				\bottomrule[\mrt]
			\end{tabular}
		\end{center}
	\end{scriptsize}
	\vspace{-5pt}
	\caption{Speed comparison with other reference methods implemented using Pytorch. Those numbers were computed using a single GTX Titan X and Intel Core i7-5820K CPU. using 320x320 inputs. Runtime statistics are computed over 200 runs.}
	\label{tab:fps_comparison}
\end{table}

\subsection{Influence of the Loss Function and the Guidance Image}
In this  section, we report  several ablation  studies to analyze  the favorable
factors of our network in  terms of performances.  Fig.~\ref{fig:pipeline} shows
the  architecture of  our  baseline  network.  All  the  following networks  are
trained using the same  setting detailed in Section~\ref{implementation details}
and trained  on the  official 795 RGB-D  images of the  NYUv2 training  set.  To
evaluate  the  effectiveness  of  our  method,  we  choose  the  predictions  of
\cite{Eigen14} as input, but note that  the conclusion is still valid with other
MDE methods. Please see supplementary material for further results.

\vspace{-3pt}

\subsubsection{Loss Functions for Depth Prediction}
In Table~\ref{tab:loss comparison}, we show the influence of different loss functions. We apply the Pytorch official implementation of $l_1$, $l_2$, and the Huber loss. The Disparity loss supervises the network with the reciprocal of depth, the target depth $y_{target}$ is defined as $y_{target} = M / y_{original}$, where $M$ here represents the maximum of depth in the scene. As shown in Table~\ref{tab:loss comparison}, our network trained with $l_1$ loss achieves the best accuracy for the occlusion boundaries.

\begin{table}[h]
	\begin{scriptsize}	
		\begin{center}
			\ra{1.1}
			\renewcommand\tabcolsep{2.8pt}
			\begin{tabular}{@{}l
					>{\centering\arraybackslash}m{0.06\linewidth}
					>{\centering\arraybackslash}m{0.06\linewidth}
					>{\centering\arraybackslash}m{0.06\linewidth}
					>{\centering\arraybackslash}m{0.08\linewidth}
					>{\centering\arraybackslash}m{0.06\linewidth}
					>{\centering\arraybackslash}m{0.06\linewidth}
					>{\centering\arraybackslash}m{0.07\linewidth}
					>{\centering\arraybackslash}m{0.07\linewidth}
					>{\centering\arraybackslash}m{0.07\linewidth}@{}
				}
				\toprule[\mrt]
				~ & \multicolumn{4}{c}{Depth error~($\downarrow$)} & \multicolumn{3}{c}{Depth accuracy~($\uparrow$)} & 
				\multicolumn{2}{c}{OB~($\downarrow$)} \\
				\cmidrule(l{3pt}r{4pt}){2-5}
				\cmidrule(l{3pt}r{4pt}){6-8}
				\cmidrule(l{3pt}){9-10}
				Method & rel & log10 & $R_{lin}$ & $R_{log}$ & $\sigma_{1}$ & $\sigma_{2}$ & $\sigma_{3}$ & $\epsilon_{a}$ & $\epsilon_{c}$ \\
				\midrule
				Baseline~\cite{Eigen14} & 0.234 & 0.095 & 0.766 & 0.265 & 0.610 & 0.886 & 0.971 & 9.926 & 9.993 \\
				\midrule
				$l_{1}$ & \textit{0.232} & \textit{0.094} & 0.758 & \textit{0.263} & \textit{0.615} & \textit{0.889} & 0.971 & \textbf{2.168} & \textbf{8.173} \\
				$l_{2}$ & \textit{0.232} & \textit{0.094} & \textbf{0.757} & \textit{0.263} & \textit{0.615} & \textit{0.889} & 0.971 & 2.302 & 8.347 \\
				Huber & \textit{0.232} & 0.095 & 0.758 & \textit{0.263} & \textit{0.615} & \textit{0.889} & \textbf{0.972} & 2.225 & 8.282 \\
				Disparity & 0.234 & 0.095 & 0.761 & 0.264 & 0.613 & 0.888 & 0.971 & 2.312 & 8.353 \\
				\bottomrule[\mrt]
			\end{tabular}
		\end{center}
	\end{scriptsize}
	\caption{Evaluation of different loss functions for learning the displacement field. The $l_1$ norm yields the best results.}
	\label{tab:loss comparison}
\end{table}

\vspace{-10pt}

\subsubsection{Guidance image}
\label{sssec:guidance}

We explore the influence of different types of guidance image. The  features of  guidance images  are extracted using an  encoder with the same  architecture as the depth  encoder, except that Leaky  ReLU~\cite{maas2013rectifier} activations  are all  replaced by  standard ReLU~\cite{glorot2011deep}.  We perform feature fusion of guidance and depth features using skip  connections from the guidance  and depth decoder respectively at similar scales.

Table~\ref{tab:guidance comparison} shows the influence of different choices for
the guidance  image.  The edge images  are created by accumulating  the detected
edges using  a series of Canny  detector with different thresholds.  As shown in
Table~\ref{tab:guidance comparison},  using the  original RGB image  as guidance
achieves  the highest  accuracy, while  using the  image converted  to grayscale
achieves   the   lowest   accuracy,   as  information   is   lost   during   the
conversion. Using the Canny edge detector can help to alleviate this problem, as
the network  achieves better results  when switching  from gray image  to binary
edge maps.





\begin{table}[h]
	\begin{scriptsize}	
		\begin{center}
			\ra{0.9}
			\renewcommand\tabcolsep{2.8pt}
			\begin{tabular}{@{}
					P{0.17\linewidth}
					>{\centering\arraybackslash}m{0.06\linewidth}
					>{\centering\arraybackslash}m{0.06\linewidth}
					>{\centering\arraybackslash}m{0.06\linewidth}
					>{\centering\arraybackslash}m{0.08\linewidth}
					>{\centering\arraybackslash}m{0.06\linewidth}
					>{\centering\arraybackslash}m{0.06\linewidth}
					>{\centering\arraybackslash}m{0.07\linewidth}
					>{\centering\arraybackslash}m{0.07\linewidth}
					>{\centering\arraybackslash}m{0.07\linewidth}@{}
				}
				\toprule[\mrt]
				~ & \multicolumn{4}{c}{Depth error~($\downarrow$)} & \multicolumn{3}{c}{Depth accuracy~($\uparrow$)} & 
				\multicolumn{2}{c}{OB~($\downarrow$)} \\
				\cmidrule(l{3pt}r{4pt}){2-5}
				\cmidrule(l{3pt}r{4pt}){6-8}
				\cmidrule(l{3pt}){9-10}
				Method & rel & log10 & $R_{lin}$ & $R_{log}$ & $\sigma_{1}$ & $\sigma_{2}$ & $\sigma_{3}$ & $\epsilon_{a}$ & $\epsilon_{c}$ \\
				\midrule
				Baseline~\cite{Eigen14} & 0.234 & 0.095 & 0.760 & 0.265 & 0.612 & 0.886 & 0.971 & 9.936 & 9.997 \\
				\midrule
				No guidance & 0.236 & 
				0.096 & 0.771 & 0.268 & 0.608 & 0.883 & 0.969 & 6.039 & 9.832\\
				
				Gray & \textit{0.232} & \textit{0.094} & \textit{0.757} & \textit{0.263} & \textit{0.615} & \textit{0.889} & \textit{0.972} & 2.659 & 8.681 \\
				
				Binary Edges & \textit{0.232} & \textit{0.094} & \textit{0.757} & \textit{0.263} & \textit{0.615} & \textit{0.889} & \textit{0.972} & 2.466 & 8.483 \\
				
				RGB & \textit{0.232} & \textit{0.094} & 0.758 & \textit{0.263} & \textit{0.615} & \textit{0.889} & 0.971 & \textbf{2.168} & \textbf{8.173} \\
				\bottomrule[\mrt]
			\end{tabular}
		\end{center}
	\end{scriptsize}
	\caption{Evaluation   of  different   ways  of   using  the   input  image for guidance. Simply using the original color image works best.}
	\label{tab:guidance comparison}
\end{table}

%% file: conclusion.tex

We showed that by predicting a displacement field to resample depth maps, we can
significantly  improve  the  reconstruction  accuracy and  the  localization  of
occlusion boundaries of any existing  method for monocular depth prediction.  To
evaluate  our method,  we also a new  dataset of  precisely
labeled  occlusion   boundaries.   Beyond   evaluation  of   occlusion  boundary
reconstruction, this dataset should be  extremely valuable for future methods to
learn to detect more precisely occlusion boundaries.

%% file: supplementary_materials.tex

\title{Predicting Sharp and Accurate Occluding Contours in Monocular Depth Estimation using Displacement Maps \\ ~ \\ Supplementary Material}
\maketitle
\thispagestyle{empty}
\blfootnote{$^*$ Authors with equal contribution.}
\section{Architecture Details}
\input{architecture}

\section{Qualitative Results}
\input{qualitative}

\section{More NYUv2-OC++ Samples}
\input{nyuoc}

%% file: architecture.tex
\begin{figure}[H]
	\begin{center}
		\icg{0.9}{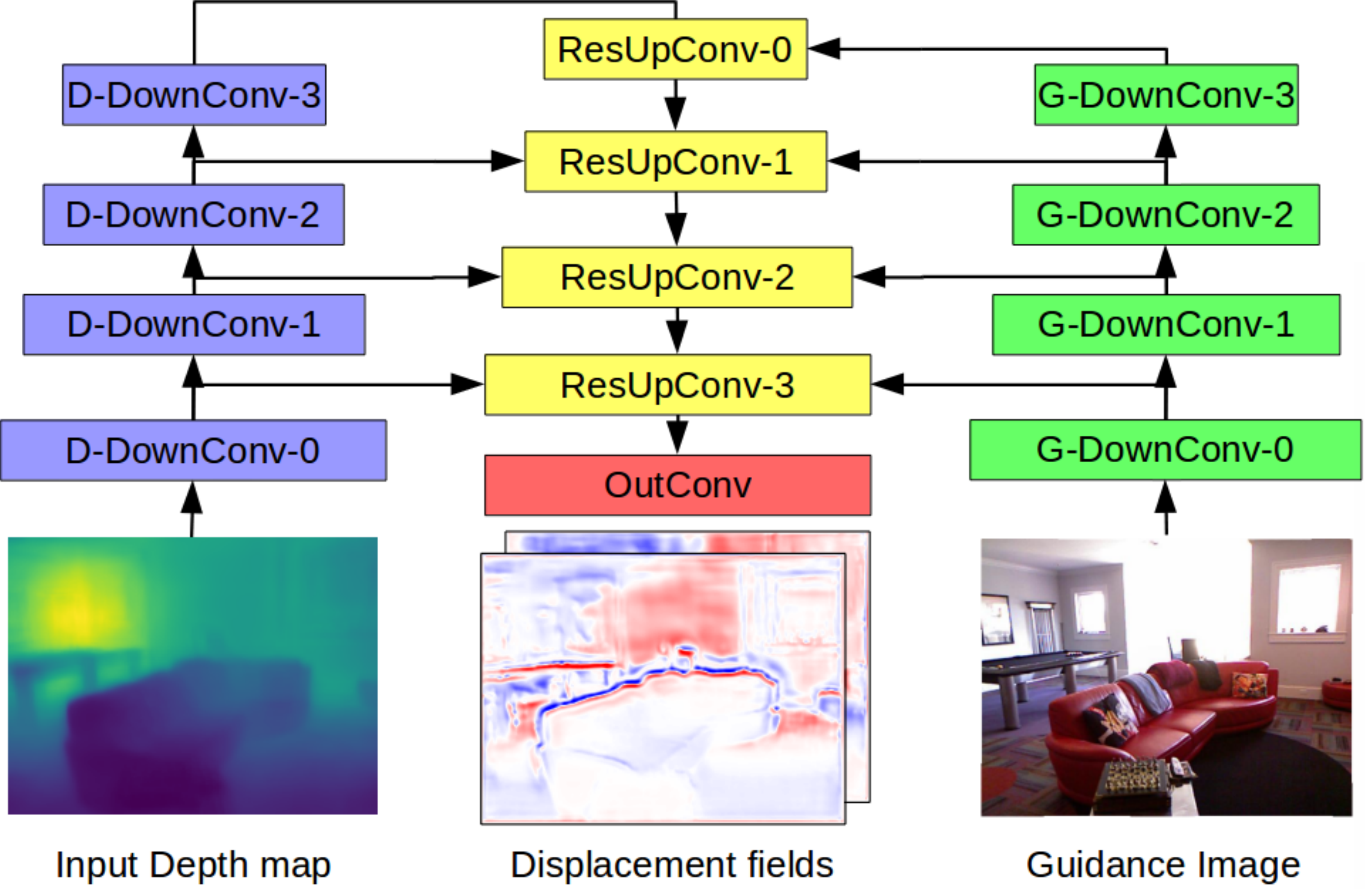}
	\end{center}		
	\caption{Detailed architecture of our displacement field prediction network. Details on the Depth and Guidance encoders are provided in Sections~\ref{ssec:depth_encoder} and~\ref{ssec:guidance_encoder} respectively.}
	\label{fig:architecture_detailed}
\end{figure}

In Fig.~\ref{fig:architecture_detailed}, we detail the architecture of our network, which is composed of two encoders, one for Depth and an optional one for Guidance. We use a single decoder which combines the respective outputs of the Depth and Guidance decoders using residual blocks and skip connections. We give full details of each block in the following.

\subsection{Depth Encoder}
\label{ssec:depth_encoder}

Our Depth is a standard encoder with a cascade of four down-convolutions, denoted \textit{D-DownConv}. The D-DownConv blocks are composed of a convolution layer with 3x3 kernel, followed by a 2x2 MaxPooling, and a LeakyReLU~\cite{maas2013rectifier} activation. The D-DownConv block convolution layers have respectively [32, 64, 128, 256] channels. They all use batch-normalization, are initialized using Xavier~\cite{Glorot10} initialization and a Leaky ReLU~\cite{maas2013rectifier} activation.

\subsection{Guidance Encoder}
\label{ssec:guidance_encoder}

Our Guidance encoder is composed of a cascade four of down-convolutions as in~\cite{He2016ResNet}, which we denote \mbox{G-DownConv}. It is identical to the D-DownConv block described in~\ref{ssec:depth_encoder} except that it uses simple ReLU~\cite{Glorot11} activations. and batch normalization for the convolution layers. The convolution layers have respectively [32, 64, 128, 256] channels.

\subsection{Displacement Field Decoder}

The displacement field decoder is composed of a cascade of four ResUpConv blocks detailed in Fig~\ref{sssec:resupconv}, and a convolution block OutConv.

\subsubsection{ResUpConv}
\label{sssec:resupconv}

\begin{figure}[h]
	\begin{center}
		\icg{0.9}{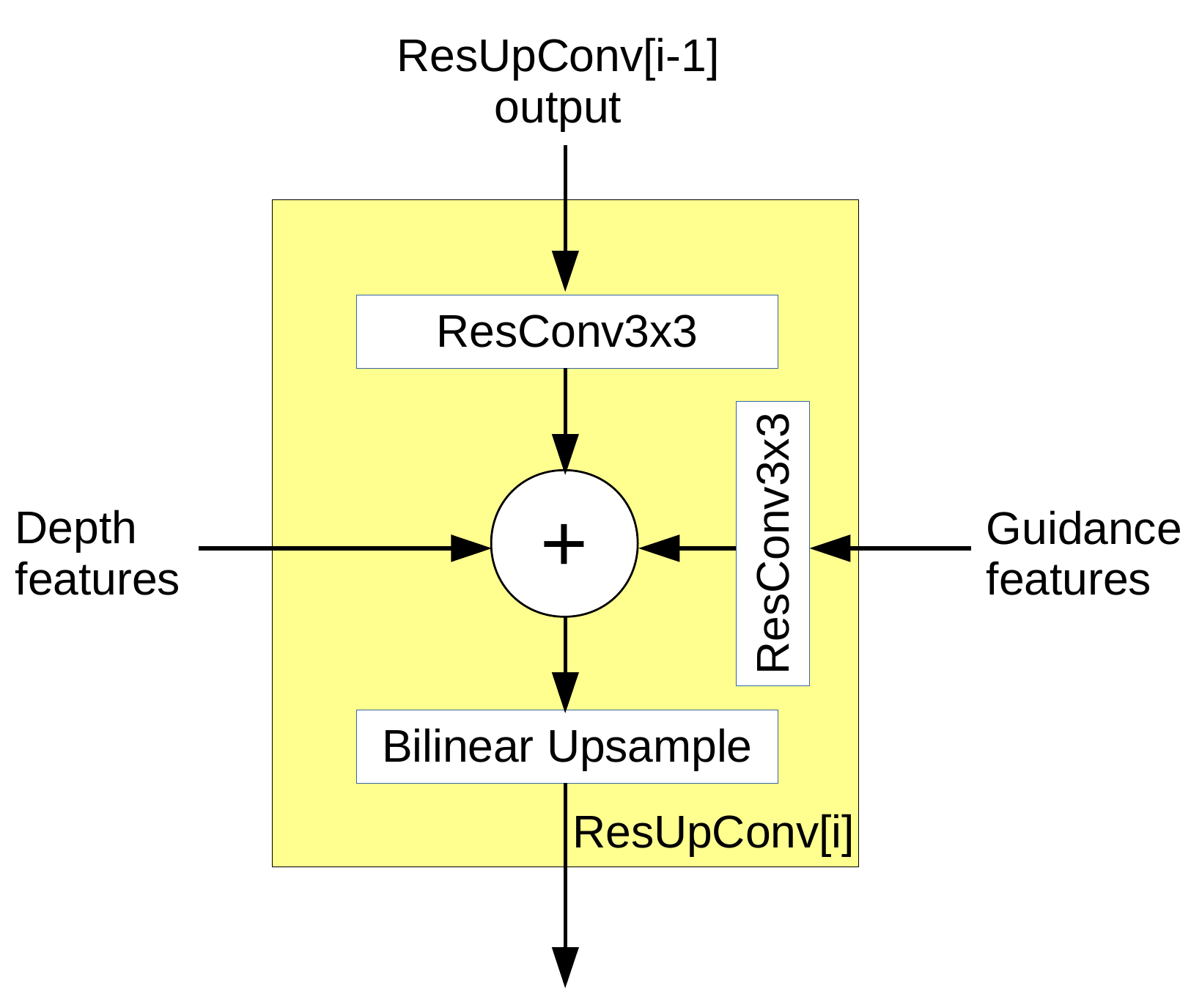}
	\end{center}		
	\caption{Details of the ResUpConv block used in our displacement field decoder.}
	\label{fig:ResUpConv}
\end{figure}

The ResUpConv block is the main component of our decoder. It fuses depth and guidance features at multiple scales using skip connections. The block architecture is detailed in Fig~\ref{fig:ResUpConv}. Guidance features are refined using a 3x3 residual convolution layer~\cite{He2016ResNet} denoted ResConv3x3. All blocks use batch-normalization, Leaky ReLU~\cite{maas2013rectifier} activation and filters weights are initialized using Xavier initialization.

\subsubsection{Output layers}
The final output block \textit{OutConv} is composed of two Conv3x3 layers with batch-normalization and ReLU~\cite{Glorot11} activation, followed by a simple 3x3 convolution layer without batch-normalization nor activation. The number of channels of those layers are respectively 32, 16, and 2. Weights are initialized using Xavier initialization.

%% file: qualitative.tex

\subsection{Comparative Results on 2D Toy Problem}
\begin{figure*}[h]
	\begin{center}
		\begin{tabular}{>{\centering\arraybackslash}m{0.05\textwidth}
				>{\centering\arraybackslash}m{0.15\textwidth}
				>{\centering\arraybackslash}m{0.15\textwidth}
				>{\centering\arraybackslash}m{0.15\textwidth}
				>{\centering\arraybackslash}m{0.15\textwidth}}
			1-a &
			\includegraphics[width=1.13\linewidth,height=1.13\linewidth]{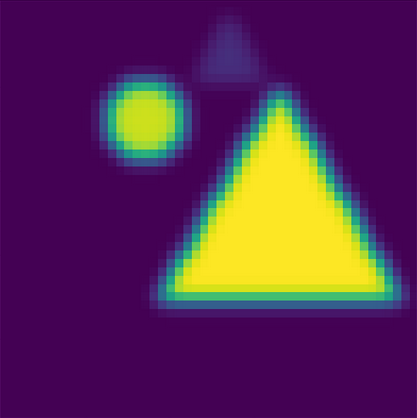} & 		 
			\includegraphics[width=1.13\linewidth,height=1.13\linewidth]{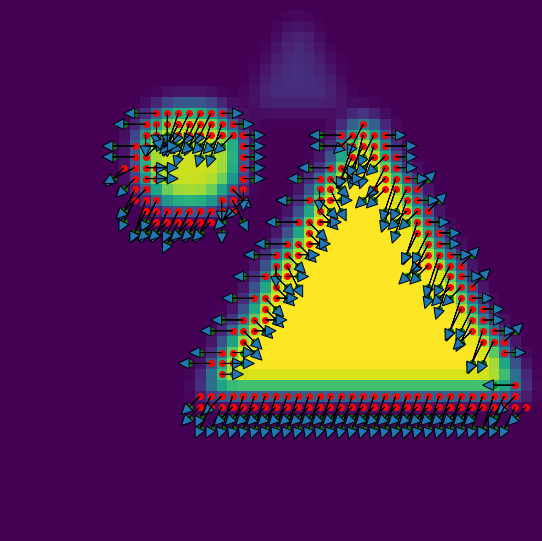} & 		 
			\includegraphics[width=1.13\linewidth,height=1.13\linewidth]{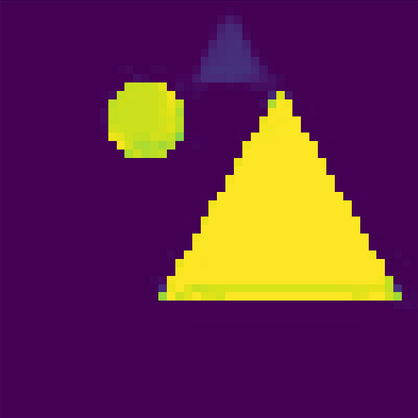} & 		
			\includegraphics[width=1.13\linewidth,height=1.13\linewidth]{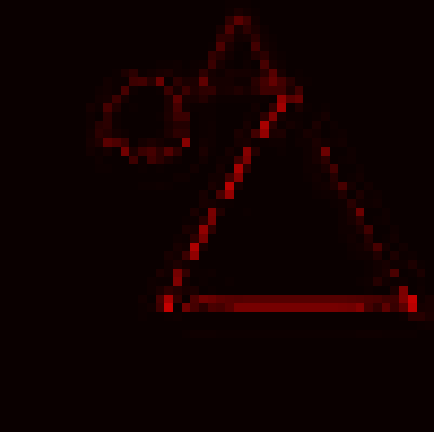}\\ 
			1-b & 
			\includegraphics[width=1.13\linewidth,height=1.13\linewidth]{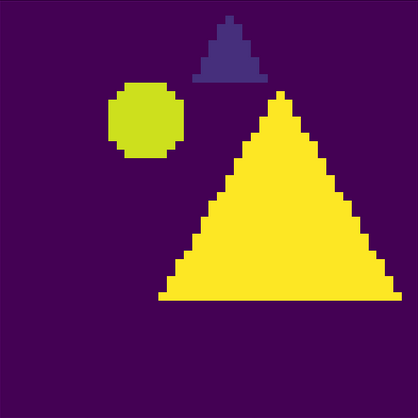} & 		 
			\includegraphics[width=1.13\linewidth,height=1.13\linewidth]{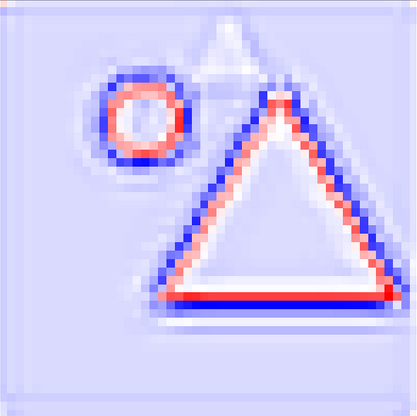} & 		 
			\includegraphics[width=1.13\linewidth,height=1.13\linewidth]{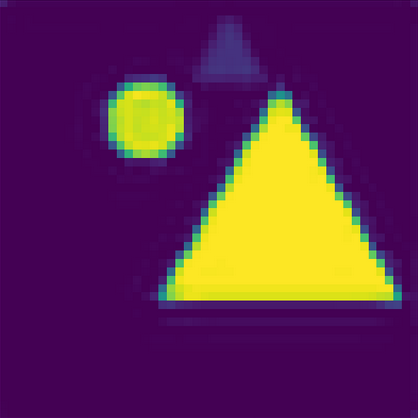} & 		
			\includegraphics[width=1.13\linewidth,height=1.13\linewidth]{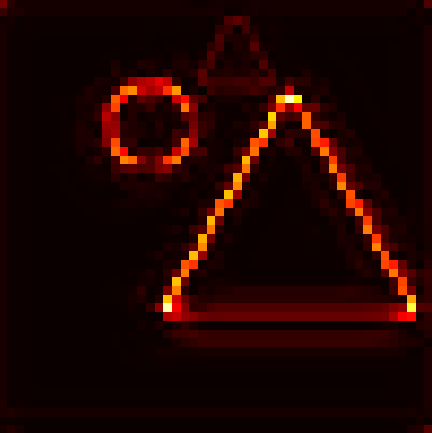}\\
			
			~ & ~ & ~ & ~ & ~ \\
			~ & ~ & ~ & ~ & ~ \\
			
			2-a & 
			\includegraphics[width=1.13\linewidth,height=1.13\linewidth]{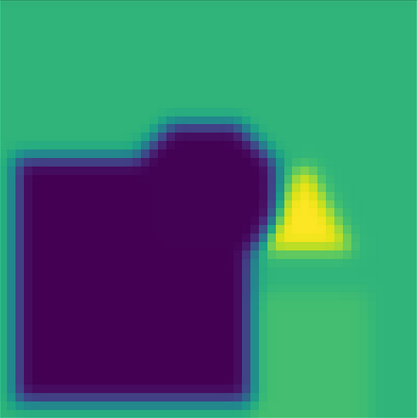} & 		 
			\includegraphics[width=1.13\linewidth,height=1.13\linewidth]{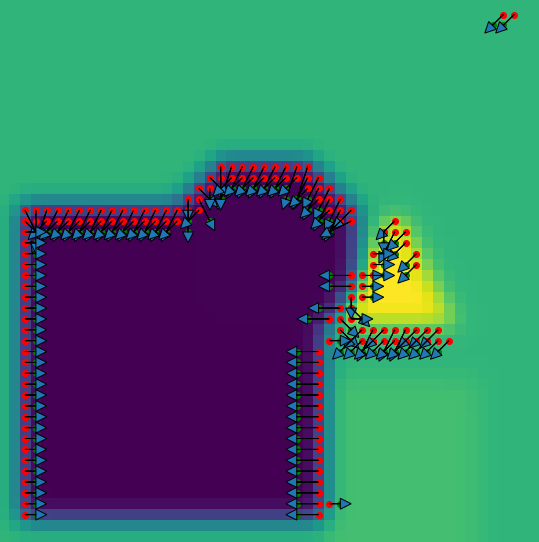} & 		 
			\includegraphics[width=1.13\linewidth,height=1.13\linewidth]{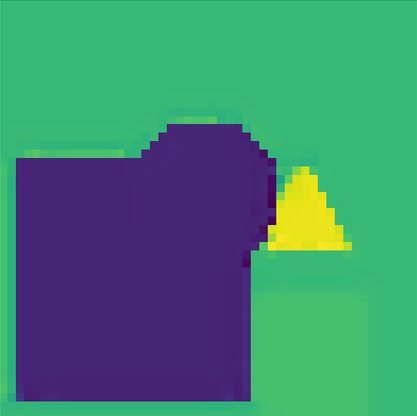} & 		
			\includegraphics[width=1.13\linewidth,height=1.13\linewidth]{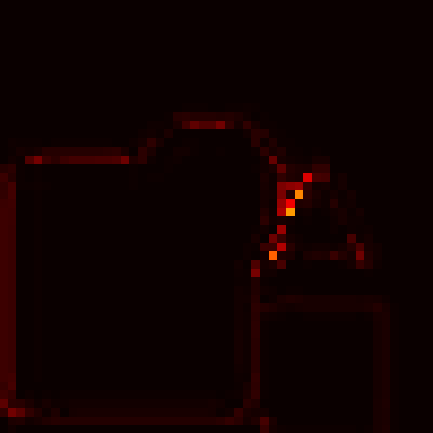}\\ 
			2-b &
			\includegraphics[width=1.13\linewidth,height=1.13\linewidth]{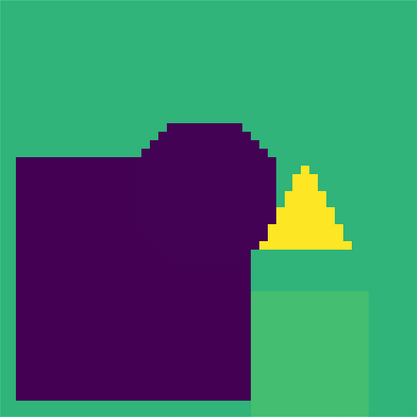} & 		 
			\includegraphics[width=1.13\linewidth,height=1.13\linewidth]{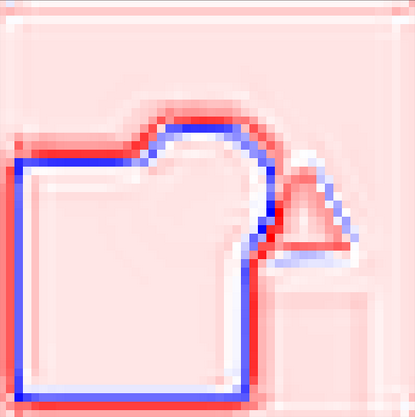} & 		 
			\includegraphics[width=1.13\linewidth,height=1.13\linewidth]{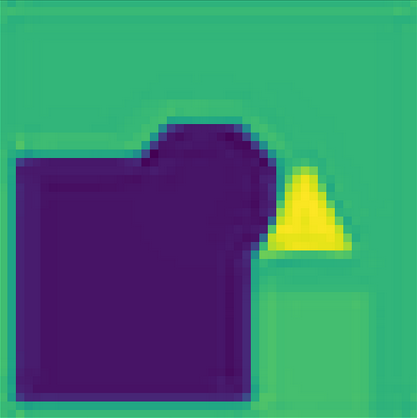} & 		
			\includegraphics[width=1.13\linewidth,height=1.13\linewidth]{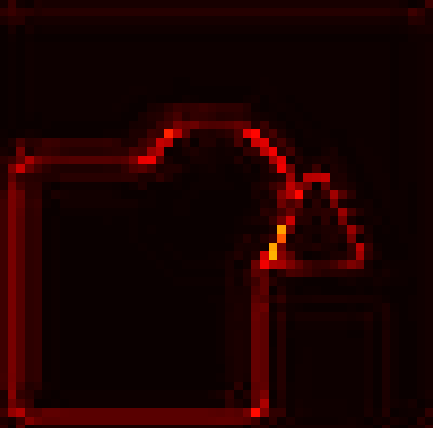}\\
			
			~ & ~ & ~ & ~ \\
			~ & ~ & ~ & ~ \\			
			
			3-a &
			\includegraphics[width=1.13\linewidth,height=1.13\linewidth]{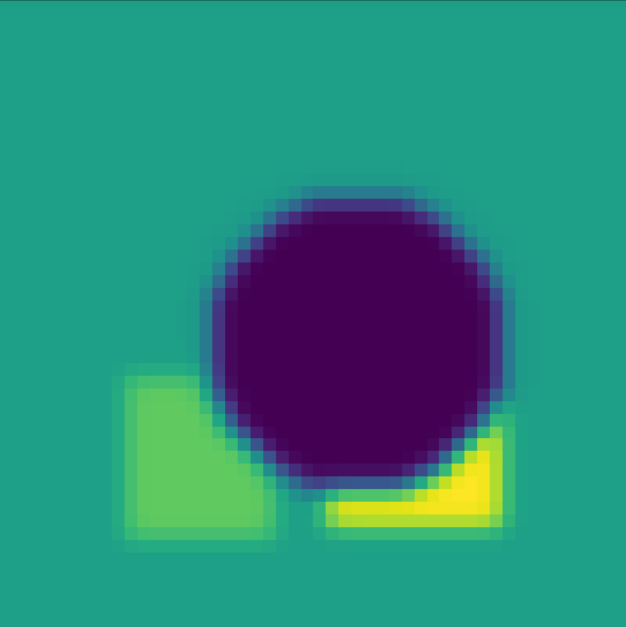} & 		 
			\includegraphics[width=1.13\linewidth,height=1.13\linewidth]{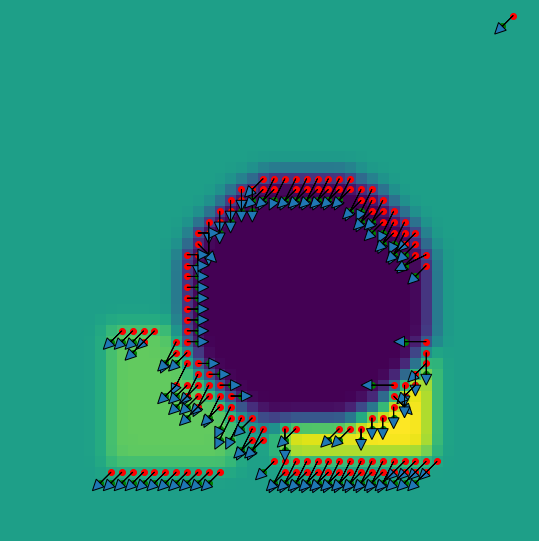} & 		 
			\includegraphics[width=1.13\linewidth,height=1.13\linewidth]{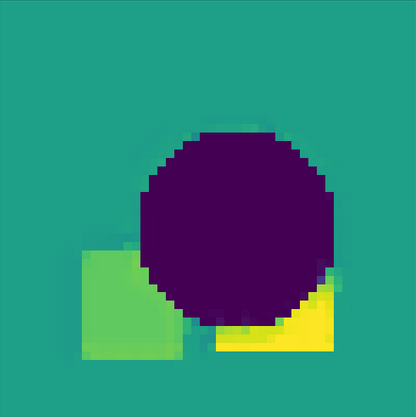} & 		
			\includegraphics[width=1.13\linewidth,height=1.13\linewidth]{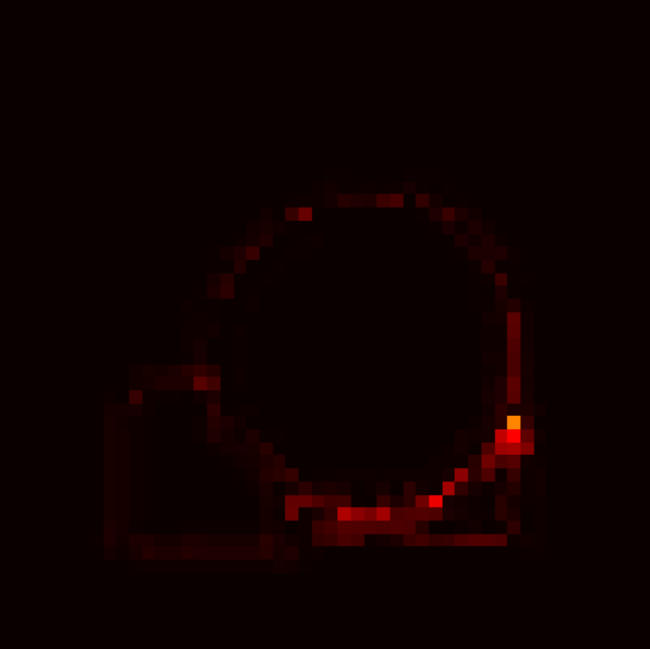}\\ 
			3-b &
			\includegraphics[width=1.13\linewidth,height=1.13\linewidth]{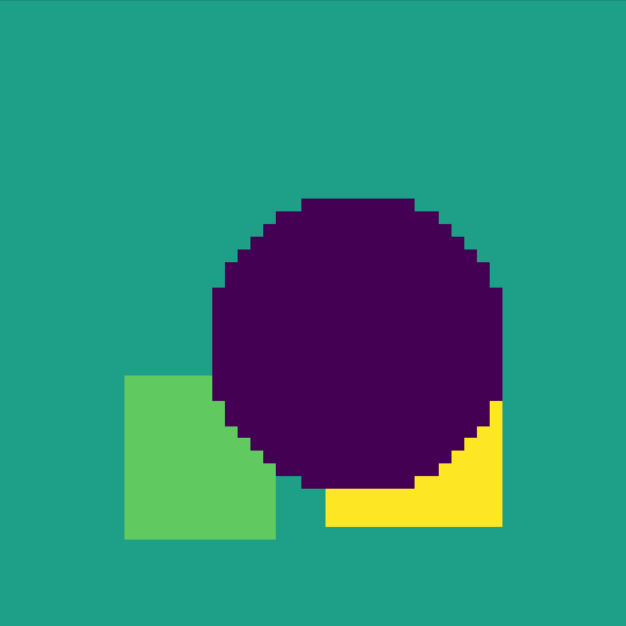} & 		 
			\includegraphics[width=1.13\linewidth,height=1.13\linewidth]{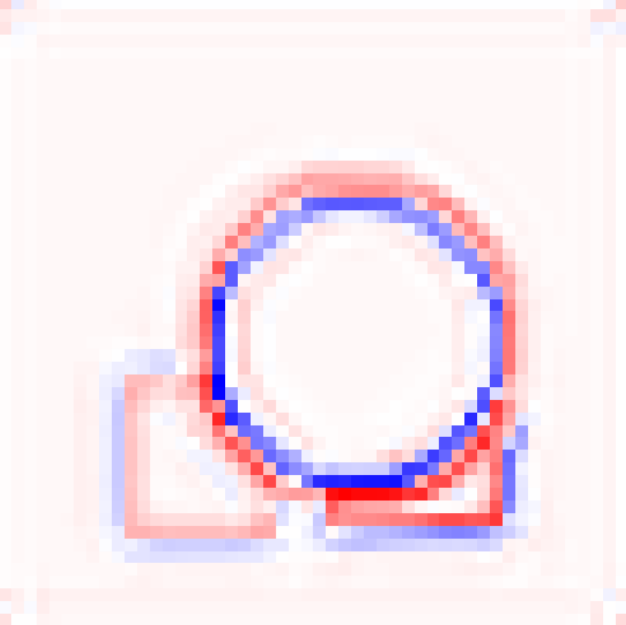} & 		 
			\includegraphics[width=1.13\linewidth,height=1.13\linewidth]{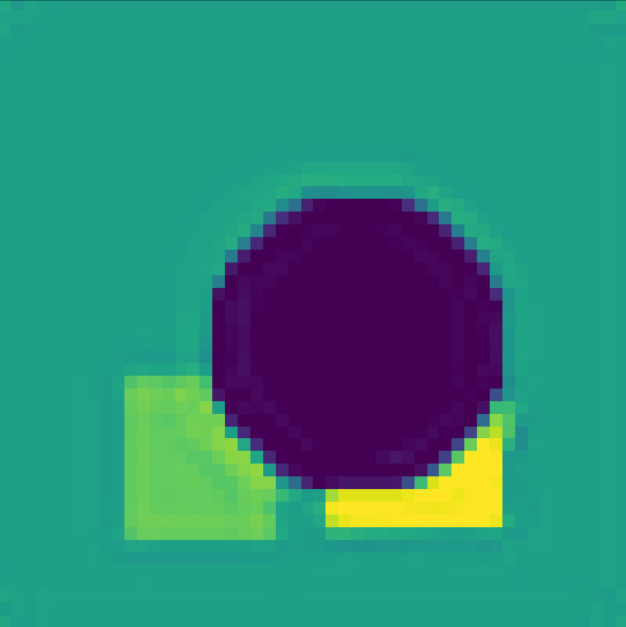} & 		
			\includegraphics[width=1.13\linewidth,height=1.13\linewidth]{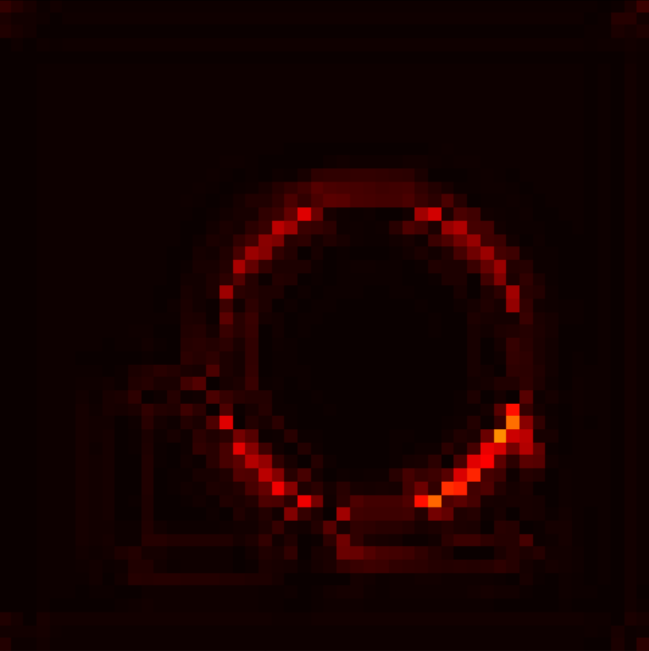}\\
		\end{tabular}
	\caption{Comparison between residual and displacement learning on a toy image sharpening problem for three examples. A blurred input image $\tilde{I}$ is fed through a Convolutional Neural Network which learns to reconstruct the original clean image $I$. Lines (1,2,3)-a show from left to right: the input image, samples of the dense predicted displacement field, refinement result with displacement update, error map. Lines (1,2,3)-b show, from left to right: the ground truth image, the predicted residual (blue is negative, red is positive, white is zero), refinement result with residual update, error map. While introducing artifacts, residual learning also results in a spread out error map around edges.}
	\label{fig:toy_problem_2D}
	\end{center}	
\end{figure*}

In Fig.~\ref{fig:toy_problem_2D}, we show qualitative results on the 2D Toy Problem described in Section \color{red} 3.3 \color{black} of the main paper. One can see that residual update introduces severe artifacts around edges, producing large and spread out errors around them. Constrastingly, our proposed displacement update recovers sharp edges without degrading the rest of the image. To ensure fair comparison between residual and displacement update methods, identical CNN architectures were used for this experiment except for the last layer which predicts a 2-channel output for the displacement field instead of the 1D-channel output residual.

\subsection{Comparative Results on NYUv2 Using \mbox{Different} MDE Methods}
\begin{figure*}[h]
	\begin{center}
	\begin{tabular}{>{\centering\arraybackslash}m{0.118\textwidth}
			>{\centering\arraybackslash}m{0.118\textwidth}
			>{\centering\arraybackslash}m{0.118\textwidth}
			>{\centering\arraybackslash}m{0.118\textwidth}
			>{\centering\arraybackslash}m{0.118\textwidth}
			>{\centering\arraybackslash}m{0.118\textwidth}
			>{\centering\arraybackslash}m{0.118\textwidth}}				
		
		\multicolumn{7}{c}{\hspace{-1em}\icg{0.99}{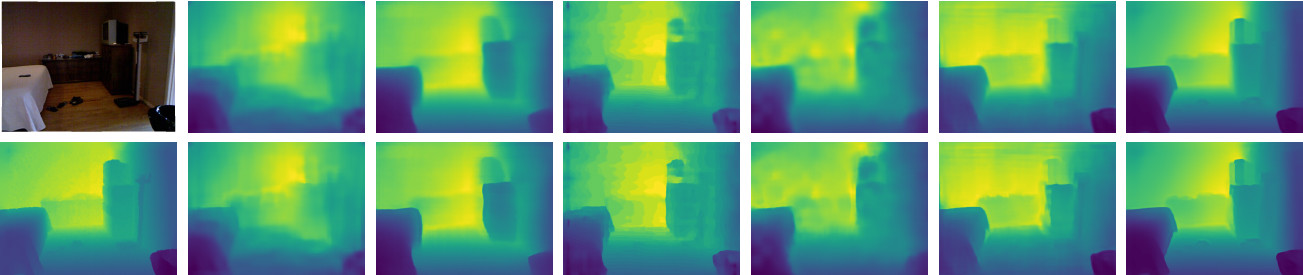}}		
		~ \\ 
		\multicolumn{7}{c}{\hspace{-1em}\icg{0.99}{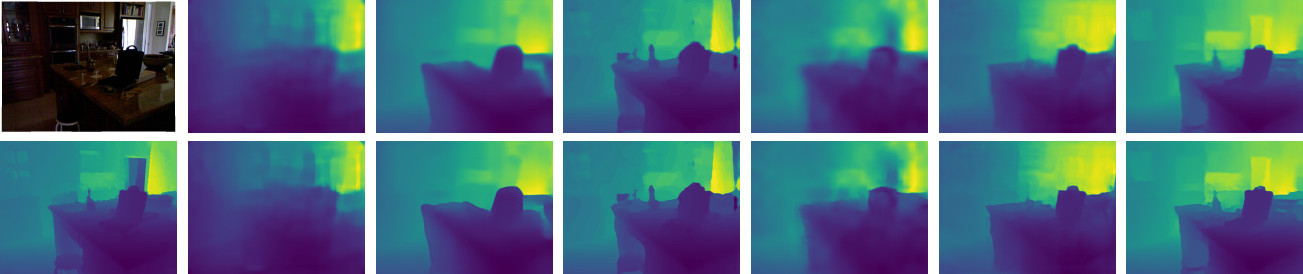}}		
		~ \\ 
		\multicolumn{7}{c}{\hspace{-1em}\icg{0.99}{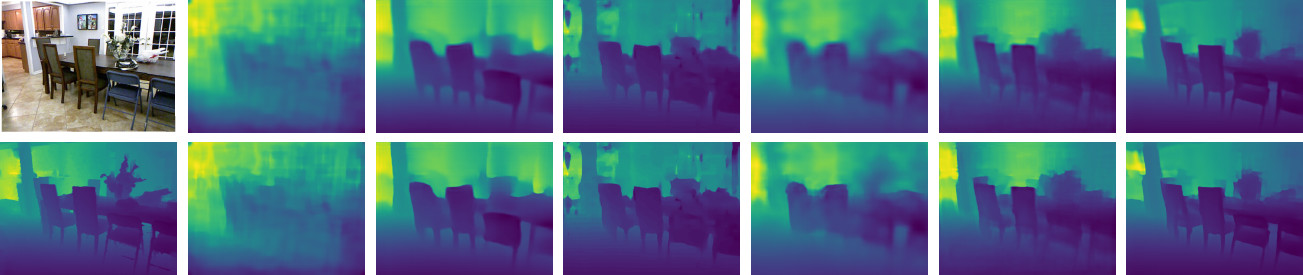}}		
		~ \\ 
		\multicolumn{7}{c}{\hspace{-1em}\icg{0.99}{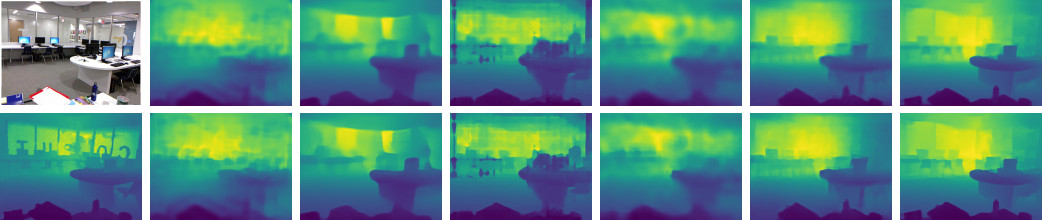}}		
		~ \\ 
		\multicolumn{7}{c}{\hspace{-1em}\icg{0.99}{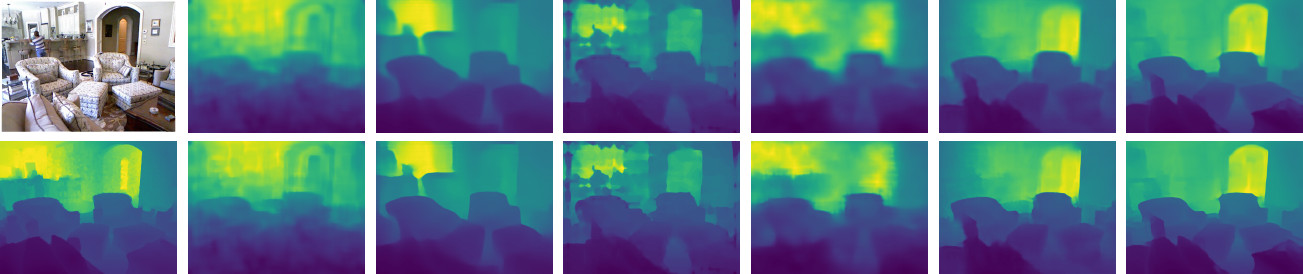}}		
		\\
		RGB image GT depth & Eigen \textit{et al.}~\cite{Eigen14} &  Laina \textit{et al.}~\cite{Laina2016DeeperDP} & Fu \textit{et al.}~\cite{FuCVPR18-DORN} & Jiao \textit{et al.}~\cite{Jiao2018LookDI}&  Ramamonjisoa \& Lepetit~\cite{SharpNet2019} &  Yin \textit{et al.}~\cite{Yin2019enforcing}
	\end{tabular}
	\caption{Refinement results using our method (best seen in color). Each example is represented on two rows, first row being the original predicted depth and second row being the refined depth. First column shows the RGB input images and associated ground truth depth from NYUv2~\cite{Nyuv2}. Following columns are refinement results for different methods we evaluated.}
	\label{fig:results}
	\end{center}
\end{figure*}

In Fig~\ref{fig:results}, we show qualitative results of our proposed refinement method on different MDE methods~\cite{Eigen14, Laina2016DeeperDP, FuCVPR18-DORN, SharpNet2019, Yin2019enforcing, Jiao2018LookDI} evaluated in the main paper. Our method always improves the sharpness of initial depth map prediction, without degrading the global depth reconstruction.

%% file: nyuoc.tex

\begin{figure*}[h]
	\begin{center}
		\begin{tabular}{>{\centering\arraybackslash}m{0.14\textwidth}
				>{\centering\arraybackslash}m{0.14\textwidth}
				>{\centering\arraybackslash}m{0.14\textwidth}
				>{\centering\arraybackslash}m{0.14\textwidth}
				>{\centering\arraybackslash}m{0.14\textwidth}
				>{\centering\arraybackslash}m{0.14\textwidth}}
			\multicolumn{6}{c}{\hspace{-1em}\icg{0.99}{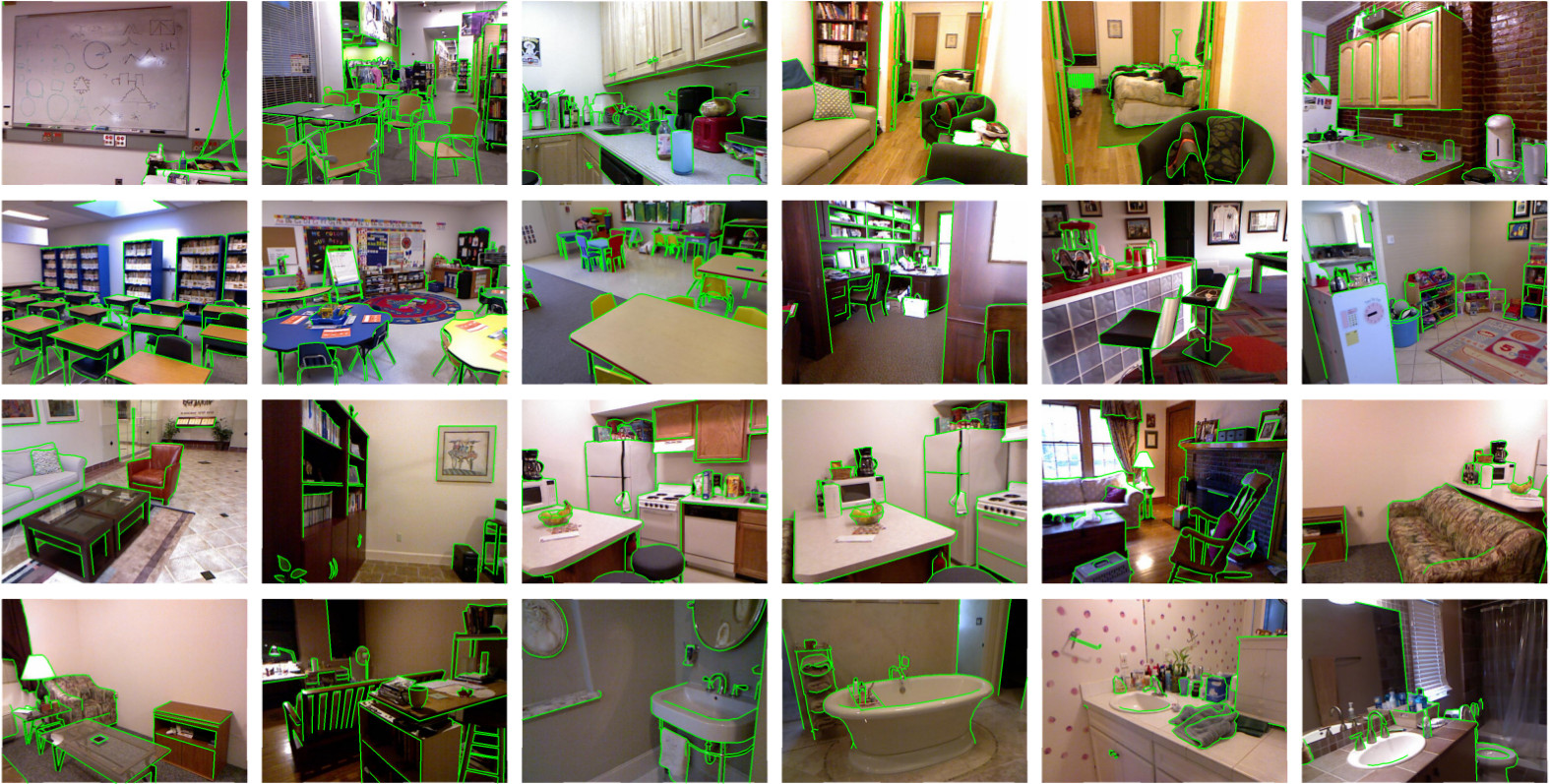}}\\ 
			\multicolumn{6}{c}{\hspace{-1em}\icg{0.99}{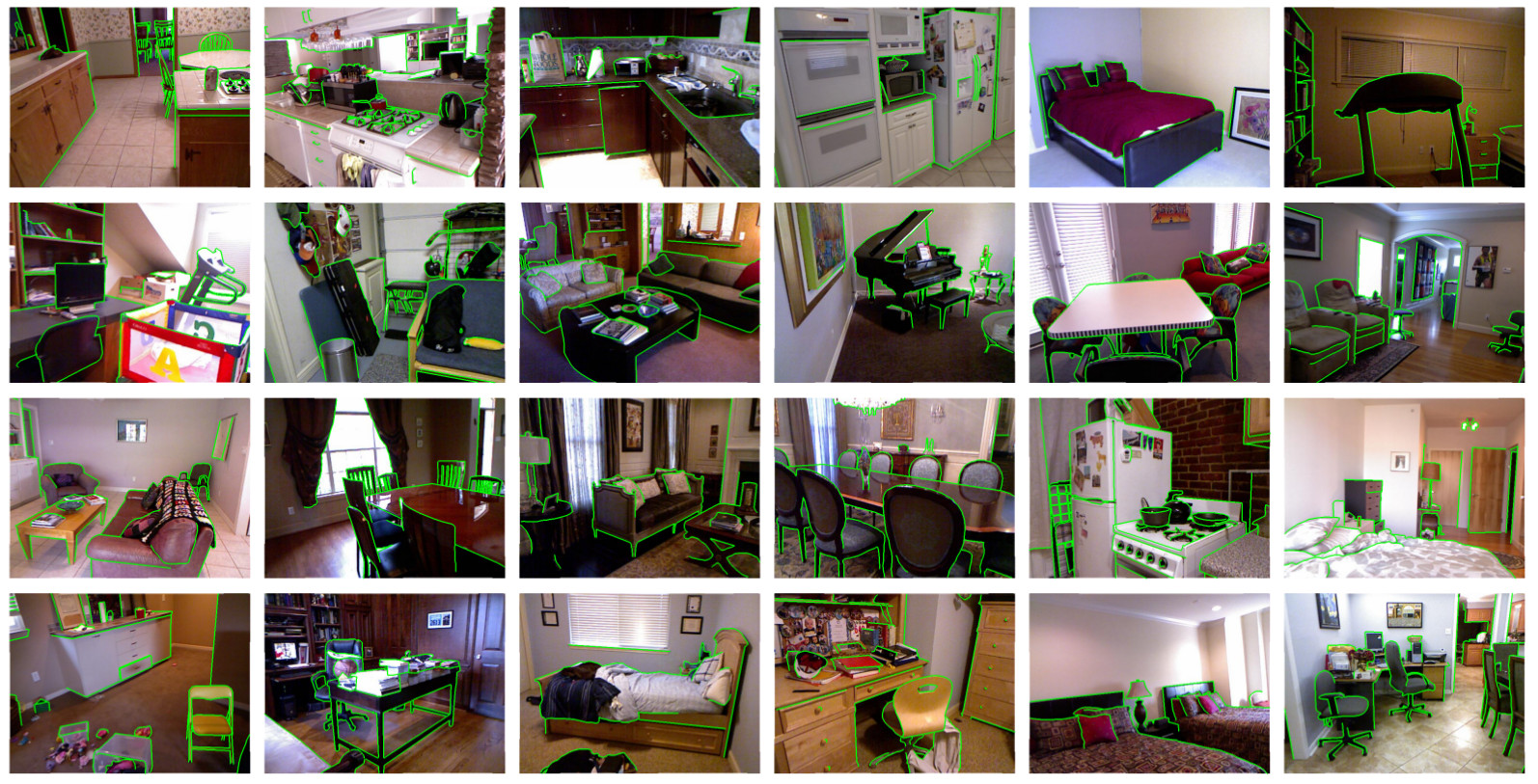}}\\ 
			\multicolumn{6}{c}{\hspace{-1em}\icg{0.99}{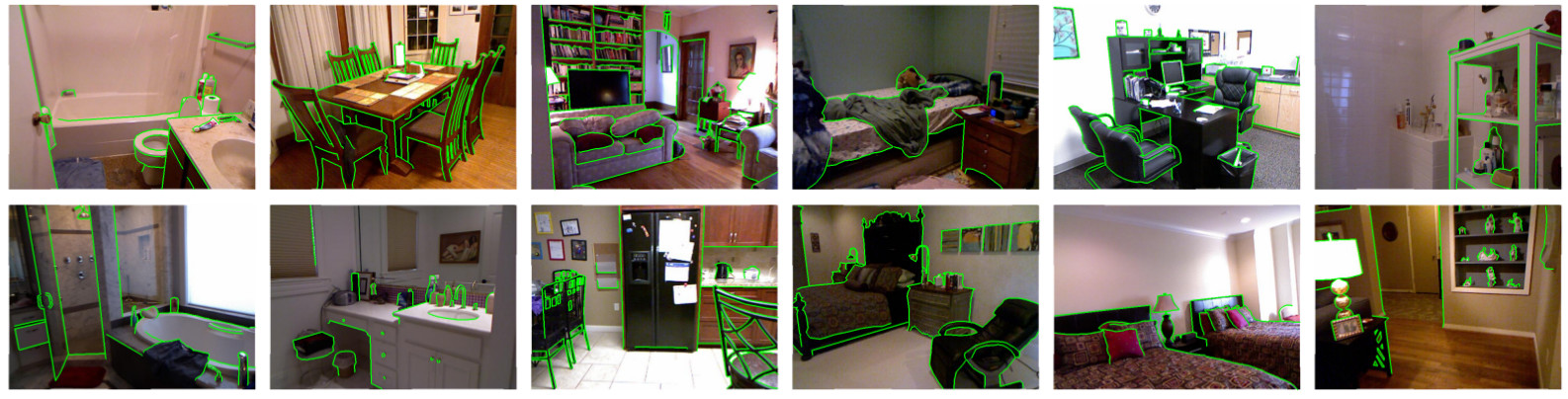}}
		\end{tabular}
		\caption{Samples taken from our fine-grained manually annotated NYUv2-OC++, which add occlusion boundaries to the popular NYUv2-Depth~\cite{Nyuv2} benchmark. We annotated the full official 654 images test set of NYUv2-Depth.}
		\label{fig:oc}
	\end{center}
\end{figure*}

In Fig.~\ref{fig:oc} we show several examples of our manually annotated 654 images NYUv2-OC++ dataset, which extends~\cite{SharpNet2019}. This dataset is based on the official test split of the popular NYUv2-Depth~\cite{Nyuv2} depth estimation benchmark.

%% file: egpaper_for_review.bbl
\begin{thebibliography}{10}\itemsep=-1pt

\bibitem{Arbelaez11}
P. Arbelaez, M. Maire, C. Fowlkes, and J. Malik.
\newblock {Contour Detection and Hierarchical Image Segmentation}.
\newblock {\em PAMI}, 33(5):898--916, 2011.

\bibitem{barron2016fast}
J.T. Barron and B. Poole.
\newblock {The Fast Bilateral Solver}.
\newblock In {\em ECCV}, 2016.

\bibitem{Canny86}
J. Canny.
\newblock {A Computational Approach to Edge Detection}.
\newblock {\em PAMI}, 8(6), 1986.

\bibitem{DengECCV18}
R. Deng, C. Shen, S. Liu, H. Wang, and X. Liu.
\newblock {Learning to Predict Crisp Boundaries}.
\newblock In {\em ECCV}, 2018.

\bibitem{Dollar13}
Piotr Doll\'ar and C.~Lawrence Zitnick.
\newblock {Structured Forests for Fast Edge Detection}.
\newblock In {\em ICCV}, 2013.

\bibitem{Eigen15}
D. Eigen and R. Fergus.
\newblock {Predicting Depth, Surface Normals and Semantic Labels with a Common
  Multi-Scale Convolutional Architecture}.
\newblock In {\em ICCV}, 2015.

\bibitem{Eigen14}
D. Eigen, C. Puhrsch, and R. Fergus.
\newblock {Depth Map Prediction from a Single Image Using a Multi-Scale Deep
  Network}.
\newblock In {\em NIPS}, 2014.

\bibitem{Everingham10}
M. Everingham, L.~Van Gool, C.K.I. Williams, J.~M. Winn, and A. Zisserman.
\newblock {The Pascal Visual Object Classes (VOC) Challenge}.
\newblock {\em IJCV}, 88(2):303--338, 2010.

\bibitem{Fan_2017_CVPR}
H. Fan, H. Su, and L.~J. Guibas.
\newblock {A Point Set Generation Network for 3D Object Reconstruction from a
  Single Image}.
\newblock In {\em CVPR}, 2017.

\bibitem{FuCVPR18-DORN}
H. Fu, A.~M. Gong, C. Wang, K. Batmanghelich, and D. Tao.
\newblock {Deep Ordinal Regression Network for Monocular Depth Estimation}.
\newblock In {\em CVPR}, 2018.

\bibitem{Fua91a}
P. Fua.
\newblock {Combining Stereo and Monocular Information to Compute Dense Depth
  Maps That Preserve Depth Discontinuities}.
\newblock In {\em IJCAI}, pages 1292--1298, August 1991.

\bibitem{Geiger2013IJRR}
A. Geiger, P. Lenz, C. Stiller, and R. Urtasun.
\newblock {Vision Meets Robotics: the KITTI Dataset}.
\newblock {\em International Journal of Robotics Research}, 2013.

\bibitem{Geiger95}
D. Geiger, B. Ladendorf, and A. Yuille.
\newblock {Occlusions and Binocular Stereo}.
\newblock {\em IJCV}, 14:211--226, 1995.

\bibitem{Glorot10}
X. Glorot and Y. Bengio.
\newblock Understanding the difficulty of training deep feedforward neural
  networks.
\newblock {\em Journal of Machine Learning Research - Proceedings Track},
  9:249--256, 01 2010.

\bibitem{glorot2011deep}
X. Glorot, A. Bordes, and Y. Bengio.
\newblock {Deep Sparse Rectifier Neural Networks}.
\newblock In {\em International Conference on Artificial Intelligence and
  Statistics}, 2011.

\bibitem{Glorot11}
X. Glorot, A. Bordes, and Y. Bengio.
\newblock {Deep Sparse Rectifier Neural Networks}.
\newblock In {\em AISTATS}, pages 315--323, 2011.

\bibitem{Monodepth17}
C. Godard, O. {Mac Aodha}, and G.~J. Brostow.
\newblock {Unsupervised Monocular Depth Estimation with Left-Right
  Consistency}.
\newblock In {\em CVPR}, 2017.

\bibitem{Gupta13}
S. {Gupta}, P. {Arbeláez}, and J. {Malik}.
\newblock {Perceptual Organization and Recognition of Indoor Scenes from RGB-D
  Images}.
\newblock In {\em CVPR}, 2013.

\bibitem{Gupta14}
S. Gupta, R. Girshick, P. Arbelaez, and J. Malik.
\newblock {Learning Rich Features from RGB-D Images for Object Detection and
  Segmentation}.
\newblock In {\em ECCV}, 2014.

\bibitem{HePAMI2013_GuidedFilter}
K. He, J. Sun, and X. Tang.
\newblock {Guided Image Filtering}.
\newblock {\em PAMI}, 35:1397--1409, 06 2013.

\bibitem{He16}
K. He, X. Zhang, S. Ren, and J. Sun.
\newblock {Deep Residual Learning for Image Recognition}.
\newblock In {\em CVPR}, 2016.

\bibitem{He2016ResNet}
K. He, X. Zhang, S. Ren, and J. Sun.
\newblock {Deep Residual Learning for Image Recognition}.
\newblock In {\em CVPR}, pages 770--778, 2016.

\bibitem{Heise2013PMHuber}
P. {Heise}, S. {Klose}, B. {Jensen}, and A. {Knoll}.
\newblock {{PM-Huber}: Patchmatch with Huber Regularization for Stereo
  Matching}.
\newblock In {\em ICCV}, 2013.

\bibitem{Heo_2018_ECCV}
M. Heo, J. Lee, K.-R. Kim, H.-U. Kim, and C.-S. Kim.
\newblock {Monocular Depth Estimation Using Whole Strip Masking and
  Reliability-Based Refinement}.
\newblock In {\em ECCV}, 2018.

\bibitem{Hoeim11}
D. Hoiem, A.A. Efros, and M. Hebert.
\newblock {Recovering Occlusion Boundaries from an Image}.
\newblock {\em IJCV}, 91(3), 2011.

\bibitem{HoeimSurfacelayout}
D. Hoiem, A.~A. Efros, and M. Hebert.
\newblock {Recovering Surface Layout from an Image}.
\newblock {\em IJCV}, 75(1):151--172, October 2007.

\bibitem{Intille94}
S.S. Intille and A.F. Bobick.
\newblock {Disparity-Space Images and Large Occlusion Stereo}.
\newblock In {\em ECCV}, 1994.

\bibitem{DaiICCV17}
H.~Qi J.~Dai, Y. Xiong, Y. Li, G. Zhang, H. Hu, and Y. Wei.
\newblock {Deformable Convolutional Networks}.
\newblock In {\em ICCV}, 2017.

\bibitem{Jaderberg15}
M. Jaderberg, K. Simonyan, A. Zisserman, and K. Kavukcuoglu.
\newblock {Spatial Transformer Networks}.
\newblock In {\em NIPS}, 2015.

\bibitem{Jeon_2018_ECCV}
J. Jeon and S. Lee.
\newblock {Reconstruction-Based Pairwise Depth Dataset for Depth Image
  Enhancement Using CNN}.
\newblock In {\em ECCV}, 2018.

\bibitem{Jiao2018LookDI}
J. Jiao, Y. Cao, Y. Song, and R.~W.~H. Lau.
\newblock {Look Deeper into Depth: Monocular Depth Estimation with Semantic
  Booster and Attention-Driven Loss}.
\newblock In {\em ECCV}, 2018.

\bibitem{Kanade94}
T. Kanade and M. Okutomi.
\newblock {A Stereo Matching Algorithm with an Adaptative Window: Theory and
  Experiment}.
\newblock {\em PAMI}, 16(9):920--932, September 1994.

\bibitem{Karsch13}
K. Karsch, Z. Liao, J. Rock, J.~T. Barron, and D. Hoiem.
\newblock {Boundary Cues for 3D Object Shape Recovery}.
\newblock In {\em CVPR}, 2013.

\bibitem{Koch2018EvaluationOC}
T. Koch, L. Liebel, F. Fraundorfer, and M. K{\"o}rner.
\newblock {Evaluation of {CNN}-Based Single-Image Depth Estimation Methods}.
\newblock In {\em ECCV}, 2018.

\bibitem{Laina2016DeeperDP}
I. Laina, C. Rupprecht, V. Belagiannis, F. Tombari, and N. Navab.
\newblock {Deeper Depth Prediction with Fully Convolutional Residual Networks}.
\newblock In {\em {3DV}}, 2016.

\bibitem{Lee2019MonocularDE}
J.-H. Lee and C.-S. Kim.
\newblock {Monocular Depth Estimation Using Relative Depth Maps}.
\newblock In {\em CVPR}, June 2019.

\bibitem{Levin2004colorization}
A. Levin, D. Lischinski, and Y. Weiss.
\newblock {Colorization Using Optimization}.
\newblock In {\em ACM Transactions on Graphics}, 2004.

\bibitem{LiICCV17}
J. Li, R. Klein, and A. Yao.
\newblock {A Two-Streamed Network for Estimating Fine-Scaled Depth Maps from
  Single RGB Images}.
\newblock In {\em ICCV}, 2017.

\bibitem{LiECCV16}
Y. Li, J.-B. Huang, N. Ahuja, and M.-H. Yang.
\newblock {Deep Joint Image Filtering}.
\newblock In {\em ECCV}, 2016.

\bibitem{Liu2018planenet}
C. Liu, J. Yang, D. Ceylan, E. Yumer, and Y. Furukawa.
\newblock {{PlaneNet}: Piece-Wise Planar Reconstruction from a Single {RGB}
  Image}.
\newblock In {\em CVPR}, 2018.

\bibitem{Depth2015CVPR}
F. Liu, C. Shen, and G. Lin.
\newblock {Deep Convolutional Neural Fields for Depth Estimation from a Single
  Image}.
\newblock In {\em CVPR}, 2015.

\bibitem{maas2013rectifier}
A.~L. Maas, A.~Y. Hannun, and A.~Y. Ng.
\newblock {Rectifier Nonlinearities Improve Neural Network Acoustic Models}.
\newblock In {\em ICML}, 2013.

\bibitem{Martin04}
D. Martin, C. Fowlkes, and J Malik.
\newblock {Learning to Detect Natural Image Boundaries Using Local Brightness,
  Color and Texture Cues}.
\newblock {\em PAMI}, 26(5), 2004.

\bibitem{Martin01}
D. Martin, C. Fowlkes, D. Tal, and J. Malik.
\newblock {A Database of Human Segmented Natural Images and Its Application to
  Evaluating Segmentation Algorithms and Measuring Ecological Statistics}.
\newblock In {\em ICCV}, 2001.

\bibitem{BerHuOwen}
B. Owen.
\newblock {A Robust Hybrid of Lasso and Ridge Regression}.
\newblock {\em Contemp. Math.}, 443, 2007.

\bibitem{paszke2017automatic}
A. Paszke, S. Gross, S. Chintala, G. Chanan, E. Yang, Z. DeVito, Z. Lin, A.
  Desmaison, L. Antiga, and A. Lerer.
\newblock {Automatic Differentiation in Pytorch}.
\newblock In {\em {NIPS Workshop}}, 2017.

\bibitem{PoggiTrinocular}
M. Poggi, A.~F. Tosi, and S. Mattoccia.
\newblock {Learning Monocular Depth Estimation with Unsupervised Trinocular
  Assumptions}.
\newblock In {\em {3DV}}, pages 324--333, 2018.

\bibitem{SharpNet2019}
M. Ramamonjisoa and V. Lepetit.
\newblock {Sharpnet: Fast and Accurate Recovery of Occluding Contours in
  Monocular Depth Estimation}.
\newblock In {\em {ICCV Workshop}}, 2019.

\bibitem{Ren12}
X. Ren and L. Bo.
\newblock {Discriminatively Trained Sparse Code Gradients for Contour
  Detection}.
\newblock In {\em NIPS}, 2012.

\bibitem{Ren2006BSDS_Ownership}
X. Ren, C.C. Fowlkes, and J. Malik.
\newblock {Figure/ground Assignment in Natural Images}.
\newblock In {\em ECCV}, 2006.

\bibitem{Unet}
O. Ronneberger, P. Fischer, and T. Brox.
\newblock {U-Net: Convolutional Networks for Biomedical Image Segmentation}.
\newblock In {\em MICCAI}, 2015.

\bibitem{Nyuv2}
N. Silberman, D. Hoiem, P. Kohli, and R. Fergus.
\newblock {Indoor Segmentation and Support Inference from RGBD Images}.
\newblock In {\em ECCV}, 2012.

\bibitem{Szeliski97b}
R. Szeliski and R. Weiss.
\newblock {Robust Shape Recovery from Occluding Contours Using a Linear
  Smoother}.
\newblock {\em IJCV}, 28(1):27--44, 1998.

\bibitem{tomasi1998bilateral}
C. Tomasi and R. Manduchi.
\newblock {Bilateral Filtering for Gray and Color Images}.
\newblock In {\em ICCV}, 1998.

\bibitem{WangSurgeNIPS16}
P. Wang, X. Shen, B. Russell, S. Cohen, B. Price, and A.~L. Yuille.
\newblock {SURGE: Surface Regularized Geometry Estimation from a Single Image}.
\newblock In {\em NIPS}, 2016.

\bibitem{Wang2016PIOD}
P. Wang and A. Yuille.
\newblock {DOC: Deep Occlusion Estimation from a Single Image}.
\newblock In {\em ECCV}, 2016.

\bibitem{wu2017fast}
H. Wu, S. Zheng, J. Zhang, and K. Huang.
\newblock {Fast End-To-End Trainable Guided Filter}.
\newblock In {\em CVPR}, 2018.

\bibitem{Xie2016Deep3DFA}
J. Xie, R.~B. Girshick, and A. Farhadi.
\newblock {Deep3D: Fully Automatic 2D-To-3d Video Conversion with Deep
  Convolutional Neural Networks}.
\newblock In {\em ECCV}, 2016.

\bibitem{xu2017multi}
D. Xu, E. Ricci, W. Ouyang, X. Wang, and N. Sebe.
\newblock {Multi-Scale Continuous CRFs as Sequential Deep Networks for
  Monocular Depth Estimation}.
\newblock In {\em CVPR}, 2017.

\bibitem{xu2018monocular}
D. Xu, E. Ricci, W. Ouyang, X. Wang, and N. Sebe.
\newblock {Monocular Depth Estimation Using Multi-Scale Continuous CRFs as
  Sequential Deep Networks}.
\newblock {\em PAMI}, 2018.

\bibitem{YangLegoCVPR19}
Z. Yang, P. Wang, Y. Wang, W. Xu, and R. Nevatia.
\newblock {LEGO: Learning Edge with Geometry All at Once by Watching Videos}.
\newblock In {\em CVPR}, 2018.

\bibitem{Yang2018UnsupervisedLO}
Z. Yang, W. Xu, L. Zhao, and R. Nevatia.
\newblock {Unsupervised Learning of Geometry from Videos with Edge-Aware
  Depth-Normal Consistency}.
\newblock In {\em AAAI}, 2018.

\bibitem{Yin2019enforcing}
W. Yin, Y. Liu, C. Shen, and Y. Yan.
\newblock {Enforcing Geometric Constraints of Virtual Normal for Depth
  Prediction}.
\newblock In {\em ICCV}, 2019.

\bibitem{ZhangPAP-CVPR19}
Z. Zhang, Z. Cui, C. Xu, Y. Yan, N. Sebe, and J. Yang.
\newblock {Pattern-Affinitive Propagation Across Depth, Surface Normal and
  Semantic Segmentation}.
\newblock In {\em CVPR}, 2019.

\bibitem{zhao2017pspnet}
H. Zhao, J. Shi, X. Qi, X. Wang, and J. Jia.
\newblock {Pyramid Scene Parsing Network}.
\newblock In {\em CVPR}, 2017.

\bibitem{BerHuZwald}
L. Zwald and S. Lambert-Lacroix.
\newblock {The Berhu Penalty and the Grouped Effect}.
\newblock 2012.

\end{thebibliography}
